\PassOptionsToPackage{table,dvipsnames}{xcolor}

\documentclass{applemlr}

\usepackage{amsmath}
\usepackage{enumerate}
\usepackage{algorithm}
\usepackage{algpseudocode}
\usepackage{amsfonts}
\usepackage{amsthm}
\usepackage{cleveref}
\usepackage{diagbox}
\usepackage{colortbl}
\usepackage{amssymb}
\usepackage{xspace}
\usepackage{wrapfig}
\usepackage{adjustbox}
\usepackage{tabularx}
\usepackage{booktabs}
\usepackage{mathtools}
\usepackage{tikz}
\usepackage{enumitem}
\usepackage{silence}
\usepackage{dsfont}
\usepackage[table]{xcolor}
\usepackage[dvipsnames]{xcolor}
\usepackage{multirow}
\usepackage{makecell}
\usepackage{xfakebold}

\usepackage{amsmath,amsfonts,bm}

\def\eqref#1{equation~\ref{#1}}

\def\1{\bm{1}}

\DeclareMathAlphabet{\mathsfit}{\encodingdefault}{\sfdefault}{m}{sl}
\SetMathAlphabet{\mathsfit}{bold}{\encodingdefault}{\sfdefault}{bx}{n}

\definecolor{textgray}{HTML}{6E6E73}
\usetikzlibrary{positioning, calc}
\usetikzlibrary{decorations.pathmorphing}

\makeatletter
\patchcmd{\wrong@fontshape}{\@gobbletwo}{}{}{}
\makeatother
\numberwithin{equation}{section}
\makeatletter
\AtBeginDocument{
  \urlstyle{sf}
  
}
\makeatother

\definecolor{light}{RGB}{125, 125, 125}
\crefname{tcb@cnt@pbox}{code}{code}
\Crefname{tcb@cnt@pbox}{Code}{Code}
\crefname{assumption}{assumption}{assumption}
\Crefname{assumption}{Assumption}{Assumptions}

\newtcolorbox[auto counter]{pbox}[2][]{
  colback=white,
  title=Code~\thetcbcounter: #2,
  #1,fonttitle=\sffamily,
  fontupper=\sffamily,
  arc=2pt,
  colframe=bgcolor,
  coltitle=fgcolor,
  colbacktitle=bgcolor,
  toptitle=0.25cm,
  bottomtitle=0.125cm
}

\makeatletter
\newcommand\applefootnote[1]{%
  \begingroup
  \renewcommand\thefootnote{}%
  \renewcommand\@makefntext[1]{\noindent##1}%
  \footnote{#1}%
  \addtocounter{footnote}{-1}%
  \endgroup
}
\makeatother

\definecolor{cverbbg}{gray}{0.90}

\counterwithout{equation}{section}

\usepackage{longtable, pdflscape}
\usepackage{siunitx}       
\usepackage{fontawesome5}  
\usepackage{xifthen}
\usepackage{float}
\usepackage{titlesec}
\usepackage{mwe}
\usepackage{dashbox}
\usepackage[textsize=tiny]{todonotes}
\usepackage[flushmargin]{footmisc}

\definecolor{backgroundcolor}{HTML}{EAF2F8} 
\definecolor{backgroundframe}{HTML}{5DADE2}
\definecolor{reasoningcolor}{HTML}{E8EAF6}  
\definecolor{reasoningframe}{HTML}{7986CB}
\definecolor{outputcolor}{HTML}{E8F5E9}     
\definecolor{outputframe}{HTML}{66BB6A}

\newtcolorbox{traceheader}{
    breakable=false,
    colback=black!3!white,
    colframe=black!20!white,
    boxrule=0.8pt, arc=3pt,
    left=6pt, right=6pt, top=4pt, bottom=4pt,
    before upper={\leavevmode}
}

\renewtcolorbox{backgroundbox}[1]{
    breakable,
    colback=backgroundcolor,
    colframe=backgroundframe,
    boxrule=1pt, arc=2pt,
    boxsep=4pt,
    before upper={\leavevmode},
    fonttitle=\bfseries\small,
    colbacktitle=backgroundframe,
    coltitle=white,
    title={#1}
}

\newtcolorbox{reasoningbox}[1]{
    breakable,
    colback=reasoningcolor,
    colframe=reasoningframe,
    boxrule=1pt, arc=2pt,
    boxsep=4pt,
    before upper={\leavevmode},
    fonttitle=\bfseries\small,
    colbacktitle=reasoningframe,
    coltitle=white,
    title={#1}
}

\newtcolorbox{outputbox}[1]{
    colback=outputcolor,
    colframe=outputframe,
    boxrule=1pt, arc=2pt,
    boxsep=4pt,
    before upper={\leavevmode},
    fonttitle=\bfseries\small,
    colbacktitle=outputframe,
    coltitle=white,
    title={#1}
}

\theoremstyle{plain}

\newcommand{\chacha}[1]{\textcolor{blue}{\textbf{[Chacha: #1]}}}

\newcommand{\matt}[1]{\textcolor{red}{\textbf{[Matt: #1]}}}
\newcommand{\nick}[1]{\textcolor{orange}{\textbf{[Nick: #1]}}}
\newcommand{\adam}[1]{\textcolor{purple}{\textbf{[Adam: #1]}}}
\newcommand{\sinead}[1]{\textcolor{pink}{\textbf{[Sinead: #1]}}}

\newcommand{\ILER}{\mathrm{I}_{\mathrm{LER}}}
\newcommand{\DKL}{D_{\mathrm{KL}}}
\newcommand{\mcX}{\mathcal{X}}

\crefname{figure}{Fig.}{Figs.}
\Crefname{figure}{Fig.}{Figs.}

\crefname{equation}{Eq.}{Eqs.}
\Crefname{equation}{Eq.}{Eqs.}

\crefname{table}{Tab.}{Tabs.}
\Crefname{table}{Tab.}{Tabs.}

\crefname{section}{Sec.}{Secs.}
\Crefname{section}{Sec.}{Secs.}

\crefname{appendix}{App.}{Apps.}
\Crefname{appendix}{App.}{Apps.}

\title{LLMs are not (consistently) Bayesian: Quantifying internal (in)consistencies of LLMs' probabilistic beliefs}

\author[*,1]{Chacha Chen}
\author[*,2,\dagger]{Matthew Jörke}
\author[1]{Adam Goli\'nski}
\author[1]{Masha Fedzechkina}
\author[1,3]{Guillermo Sapiro}
\author[1]{\mbox{Sinead Williamson}}
\author[1]{Nicholas Foti}

\affiliation[1]{Apple}
\affiliation[2]{Stanford University}
\affiliation[3]{Princeton University}

\abstract{Modern AI systems are being deployed in complex domains such as medicine, science, and law, where there is often not a single correct answer given the observed evidence. Such systems must be able to represent and update uncertain beliefs about the world as new evidence arrives to make rational decisions.
We introduce the novel technique of studying LLMs as information processing rules and utilize the \textit{information processing gap} -- the deviation from Bayes updates -- to study the internal (in)consistencies of how LLMs update their probabilistic beliefs from evidence. 
Our extensive experiments evaluate multiple approaches in which LLMs can incorporate evidence into their beliefs.
Some of these approaches produce (nearly) Bayesian updates, thus optimally processing evidence; others use a learned heuristic.
Surprisingly, the non-Bayesian heuristic updates often outperform exact Bayesian updates (optimal information processing) in terms of downstream task performance---indicating the LLMs' probabilistic models of the world are misspecified. 
Lastly, we show how our measure can provide
diagnostics to identify issues with LLM-powered inferential systems.

}

\metadata[Correspondence]{\sffamily Nicholas Foti: \url{nicholas_foti@apple.com}}
\date{\sffamily\today}

\begin{document}

\maketitle

\applefootnote{\textsuperscript{*}Equal contribution, random author order. \textsuperscript{$\dagger$}Work done while an intern at Apple.}

\section{Introduction}

The impressive capabilities of large language models (LLMs) has lead them to become core inference- and decision-making engines for systems in high-stakes problem settings such as medical diagnosis and constructing legal arguments \citep{papamarkou2026agentic,brodeur2026performance,dehghani2025large}.
In these settings, the LLM is provided with evidence -- e.g., a patient's symptoms -- and is expected to update its beliefs about the patient's diagnosis based on those symptoms.
The implicit hope is that LLMs perform this mapping from evidence to beliefs in a principled and rational manner.
Since it is well established that Bayes' Theorem is the decision-theoretically optimal way to update beliefs based on new evidence \citep{Robert1994Bayesian}, existing work has been focused on understanding whether LLMs ``are Bayesian''~\citep{liu2026ministral,aliakbar2025,schrader2024}, or encouraging them to be ``more Bayesian'' via prompting or relative to an oracle~\citep{nori2025sequential,shaikh2025creating,qiu2026bayesian}.

However, there has been little work studying and quantifying how LLMs incorporate evidence, in the absence of a clearly specified statistical model. 
Understanding how LLMs update their beliefs from evidence is imperative for trusting, auditing, and interpreting the behaviors of decision systems built on LLMs that must reason from empirical evidence, and anticipating how the LLM will update its beliefs in previously unseen settings.
This work takes a fundamental step in understanding how LLMs incorporate evidence to update their beliefs about uncertain outcomes.
We approach this from a information-theoretic angle, treating LLMs as \textit{information processing rules} and measuring the \textit{information processing gap} between the LLM's resulting post-data beliefs $q$ and the information it thinks is in the evidence (the theoretical justification for this metric is introduced later). This information processing gap $\Delta(q)$ is minimized under exact Bayesian updating behavior, as will be later shown in this work, and allows us to measure how far from Bayesian a model is without depending on a Bayesian oracle or training data. Moreover, this information processing gap can be decomposed into interpretable components that allow us to analyze the nature of any deviations from Bayesian updating.

We perform empirical evaluation, with multiple open and closed source LLMs, on a series of multiple-choice question-answering tasks where partial evidence is revealed sequentially, an easily specified task that contains many of the core aspects of sequential inference. Through our experiments, we find:
\begin{itemize}[leftmargin=*]
    \item LLM's belief updating mechanism differs depending on whether it is asked to explicitly update its beliefs after each evidence step -- belief propagation (BP) mode -- vs.\ if it is asked to incorporate multiple pieces of evidence simultaneously (batch mode) in two respects.
    First, on average, the beliefs obtained using batch mode yield better task performance. Second, LLMs are consistently closer to Bayesian updating in BP mode than in batch mode.
    \item Forcing an LLM to explicitly perform Bayes updates does not consistently lead to better task performance. Further, when the LLM's post-data beliefs disagree with its likelihood model, they are often aligned with an ``oracle'' likelihood from a stronger LLM. 
    \item We find strong (Spearman's) correlation between batch-mode information processing gap and BP task performance, providing a diagnostic check of when one can trust the BP mode inferences.
\end{itemize}

{\bf Key contribution:} This paper introduces the novel study of LLMs as information processing rules, formalizing this concept via the optimality relationship between information processing and Bayesian updates, and provides extensive empirical findings and unintuitive lessons about how LLMs process information into beliefs when presented with new evidence.
\section{Measuring Information Processing Capabilities}
\label{sec:zellner}
Our analysis of how LLMs update their beliefs from evidence is based on an information theoretic approach~\citep{zellner1988optimal}. We first define the necessary probabilistic quantities and terms and then present our proposed measure.
Let $\theta \in \Theta$ denote a hypothesis space -- e.g., medical conditions that a patient may have -- and let $X \in \mathcal{X}$ denote the space of observed evidence -- e.g., symptoms the patient exhibits.
We consider sequences of evidence $X_{1:n} \triangleq (X_1,\ldots,X_n) \in \mcX^n$ presented over $n \in \mathbb{N}$ \textit{evidence steps}.

Suppose that \textit{prior evidence} $X_{1:n}$ is given, we observe \textit{new evidence} $X_{n+1}$, and wish to update our \textit{belief} (i.e., probability distribution) over hypotheses $\theta \in \Theta$. We define the following quantities:
\begin{itemize}[leftmargin=*]
    \item $\pi(\theta | X_{1:n})$ is the \textit{prior} distribution over hypotheses given previously collected evidence.
    
    \item $\ell(X_{n+1} | \theta, X_{1:n}) : \Theta \to [0,1]$ is the \textit{likelihood} of $X_{n+1}$.
    
    \item $q(\theta | X_{1:n+1})$ denotes a \textit{post-data} distribution over hypotheses given the prior and new evidence.

    \item $p(\theta | X_{1:n+1}) \propto \pi(\theta | X_{1:n}) \ell(X_{n+1}| \theta, X_{1:n})$ denotes the posterior computed from a particular pair of prior $\pi$ and likelihood $\ell$ via an explicit application of Bayes' Theorem.
    
\end{itemize}
We emphasize that we do \textbf{not} assume that $\pi$, $\ell$, and $q$ are induced by an underlying joint distribution. Rather, $\pi$, $\ell$, and $q$ are treated as free variables. We use the term \emph{post-data} instead of posterior to emphasize that $q$ is not assumed to be determined by Bayes' Theorem. 
In the experiments, 
we will elicit $\pi$, $\ell$, and $q$ from LLMs.

The information in a distribution $d$ is a key quantity and is defined as $\mathrm{I}[d] \triangleq \mathbb{E}_{q(\theta | X_{1:n+1})} [ \log d(\theta)]$. 
This expectation is with respect to the post-data distribution, and corresponds to the negative (cross-) entropy for $\pi$ and $q$, 
such that a lack of surprise (i.e., lower entropy) corresponds to more information.

An \textit{information processing rule} is defined as a mapping from inputs $(\pi, \ell)$ to output $q$. 
We define \textit{input information} $\mathrm{I}_{\mathrm{in}}$, \textit{output information} $\mathrm{I}_{\mathrm{out}}$, and the \textit{information processing gap} $\Delta(q)$ as
\begin{equation}
    \textstyle \mathrm{I}_{\mathrm{in}} \triangleq \mathrm{I}[\pi] + \mathrm{I}[\ell],
    \quad
    \mathrm{I}_{\mathrm{out}} 
    \triangleq \mathrm{I}[q] + \mathrm{I}[Z_{\pi,\ell}],
    \quad
    \Delta(q) \triangleq \mathrm{I}_{\mathrm{out}} - \mathrm{I}_{\mathrm{in}}
    = \DKL(q || p) \ge 0,
    \label{eq:zellner-delta}
\end{equation}
where 
$Z_{\pi,\ell} = \mathbb{E}_{\pi(\theta | X_{1:n})}[\ell(X_{n+1}|\theta, X_{1:n})]$ is the marginal likelihood under the prior $\pi$ and hence is no longer a function of $\theta$,
and 
$p$ is the posterior computed by Bayes' Theorem using $\pi$ and $\ell$. 

The information processing gap quantifies an information processing rule's efficiency. If $\Delta(q) > 0$, the post-data distribution is inconsistent with the prior and the likelihood. 
Alternatively, $\Delta(q) = 0$ implies optimal information processing and occurs only when the post-data distribution is obtained via Bayes' Theorem~\citep{zellner1988optimal}.
Thus, $\Delta(q)$ can provide diagnostic information on the behavior of LLMs, where larger values indicate a greater divergence from optimal information processing. 

While $\Delta(q)$ is not a new metric, and is connected to the ELBO used in variational inference \citep{Blei2017,alemi2020a,mukherjee2019}, it has never needed to be used as a model diagnostic. Traditional inference algorithms are designed to be  (approximately) optimal information processing rules. However, inference is an emergent property in LLMs, and has not been ``designed'', and $\Delta(q)$ is a natural quantity to measure this emergent capability of LLMs.

\subsection{Decomposing the information processing gap}\label{sec:decomposition}
While $\Delta(q)$ measures the extent to which an information processing rule is suboptimal, one can also examine its components to pinpoint \textit{why} it is suboptimal.
To this end, we rearrange \Cref{eq:zellner-delta} into more interpretable terms involving the pre- to post-data KL divergence and expected log likelihood-evidence-ratio (abbreviated as LER),
\begin{equation}
    \Delta(q) = \mathrm{I}_{\mathrm{out}} - \mathrm{I}_{\mathrm{in}} \equiv \DKL(q||\pi) - \ILER,
    \label{eq:kl-ler}
\end{equation}
where $\DKL(q||\pi) \triangleq \mathrm{I}[q] - \mathrm{I}[\pi]$ is the KL-divergence from the post-data distribution to the prior, and
\begin{align}
    \ILER 
    &\textstyle \triangleq 
    \mathrm{I}[\ell] - \mathrm{I}[Z_{\pi,\ell}]
    = 
    \mathbb{E}_{q(\theta | X_{1:n+1})}\left[
    \log 
    \frac{\ell(X_{n+1} | \theta, X_{1:n})}{Z_{\pi,\ell}}
    \right].
    \label{eq:Iler}
\end{align}
This decomposition exposes a structural property of optimal information processing, in which the KL divergence from the pre- to post-data beliefs, $\DKL(q||\pi)$, 
should equal the expected likelihood evidence ratio, $\ILER$. 
$\DKL(q||\pi)$ measures how the beliefs have changed under the update. 
$\ILER$ looks at how much the post-data distribution is aligned with the new evidence ($\mathbb{E}_q[\log \ell(X_{n+1}|\theta, X_{1:n})]$), relative to the log marginal likelihood under the prior, $\log Z_{\pi,\ell}$. 
Loosely, $\ILER$ tells us how much of the model's update is justified by the evidence. 
The information gap $\Delta(q)$ then can be interpreted as the ``unjustified'' component of the total update, capturing the intuition that the amount we change our beliefs in the direction of a hypothesis from the prior should align with how strong the evidence supports the hypothesis. 
This is intuitive to a Bayesian as it is exactly what Bayes' Theorem does. However, in the age of LLMs, \Cref{eq:kl-ler,eq:Iler} provide novel insights into the underlying processing, since the models are not explicitly trained to follow Bayes' Theorem.

There are a number of ways in which an update might deviate from Bayes' Theorem.  It might move in the ``wrong direction,'' which we define as when the new evidence $X_{n+1}$ has lower expected log-likelihood under the post-data distribution 
than under the prior, 
i.e., $\mathbb{E}_q[\log \ell] < \mathbb{E}_\pi[\log \ell]$, implying that $\ILER < -\DKL(\pi || p)$. In this setting, probability mass has shifted towards hypotheses the evidence does not support.

Alternatively, updates may be in the ``right direction,'' which \Cref{eq:kl-ler} helps to characterize. Since $\ILER$ is monotonically increasing as we move from the prior in the direction of the Bayesian posterior, any update that maintains the Bayesian direction satisfies $\ILER \geq -\DKL(\pi || p)$. 
In this regime, we may see under-updating ($\DKL(q || \pi) < \DKL(p || \pi)$), where the distribution moves in the right direction but not far enough, or  over-updating ($\DKL(q || \pi) > \DKL(p || \pi)$)). 
Note that while extreme over-updates can produce a large $\Delta(q)$,      
indicating substantial deviation from optimal information processing, this is qualitatively different from wrong-direction updates.

\section{Methodology}

We carry out experiments on four standard datasets of multiple-choice question-answering tasks, where the model must assign beliefs to options given partial evidence. 
We break up each data point into a sequence of \textit{evidence steps} where at each step we reveal a new piece of evidence to the LLM. We construct four datasets (details in \Cref{app:datasets}): \textit{Animals}, where the task is to identify an animal based on its attributes; \textit{Political ideology}, where the task is to identify a respondent's political ideology from their responses to survey questions; \textit{MediQ}, where the task is to diagnose a patient based on evidence from a clinical visit; and \textit{Eleusis}, where the task is to identify a hidden rule in a card game, based on cards played.

At each step we elicit post-data distributions and likelihoods for each question by prompting the LLM to ``verbalize'' probablities in the token space, as introduced by \citet{tian2023just}. 
We consider two strategies for extracting the LLM's post-data distribution after a given evidence step. 
 The first we call \textit{belief propagation (BP) mode}. 
After each piece of evidence, the LLM is provided its elicited likelihood as part of the prompt, and it is prompted to explicitly update its beliefs. 
Crucially, we do not present the previous evidence steps. This mirrors the realistic scenario where an LLM must repeatedly update its beliefs during multiple interactions with a user.  
The second, we call \textit{batch mode}. Here, we present all the evidence up to the current step in a single prompt, and ask the LLM to provide a distribution over the hypothesis space. We do not provide any intermediate beliefs or likelihoods. This mirrors the scenario where the model must incorporate multiple pieces of evidence at once, taking into account their ordering and any dependencies. 
See \Cref{fig:prompt-templates-2,fig:prompt-placeholder-values} for details.

From these distributions we can compute the information gap $\Delta(q)$. We also consider two measures of the utility of the distributions in downstream tasks (described in more detail in \Cref{app:metrics}): Area under the receiver operating characteristic curve (AUROC), and expected calibration error (ECE). We evaluate on multiple LLMs, both open-source models (Qwen3-\{8B, 14B\} \citep{yang2025qwen3}, Ministral-\{8B, 14B\} (reasoning and non-reasoning versions) \citep{liu2026ministral}, DeepSeek-R1-0528-Qwen3-8B \citep{guo2025deepseek}); and closed-source models (GPT-4o mini, GPT-5.1 with both medium and no reasoning effort).

\section{Results}
\label{sec:results}

Figures in the main text show the largest evidence step such that $\geq 80\%$ of the observations had valid post-data distributions and likelihood values returned for all models analyzed (see \Cref{tab:dataset_statistics,fig:n_questions_trajectories}). This filtering is necessary since sometimes models fail to respond or they return non-parseable output. Additional evidence steps are included in the Appendix. 
Where appropriate, $\rho$ indicates Spearman's rank correlation. All 95\% error bars are calculated via bootstrapping.

\subsection{Better task performance and calibration in batch mode than in BP mode}\label{sec:task_performance}

\begin{figure*}[ht]
    \centering
    \includegraphics[width=0.9\linewidth]{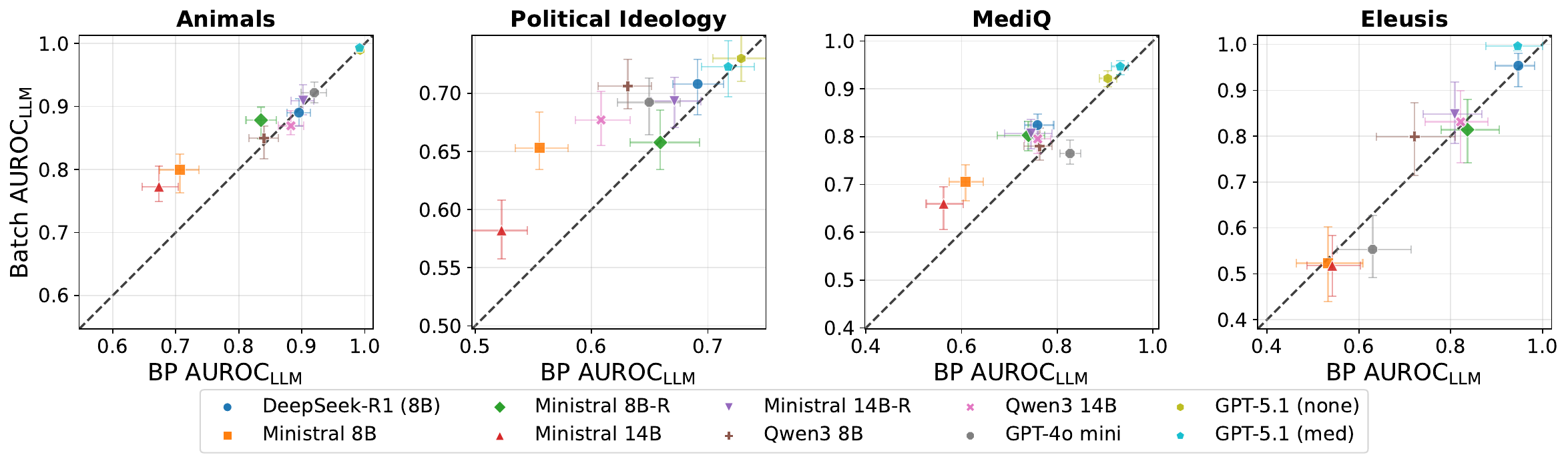}
    \caption{
    Comparing task performance (AUROC) between batch and BP modes. Equal performance is indicated by the dotted line. Note how in general, batch outperforms BP.
    }
    \label{fig:batch_vs_seq_auroc}
\end{figure*}

We consider the task performance of each model at the largest evidence step determined for each dataset. 
\Cref{fig:batch_vs_seq_auroc} shows that weaker models attain higher task performance in batch mode than in BP mode on the Animals, Political Ideology, and MediQ datasets. 
More capable models perform comparably in both modes. We also verify that as LLMs see more evidence they achieve better task performance (see \Cref{fig:auroc_evidence_step}).
We do not observe noticeable differences in task performance between modes on Eleusis. 
This dataset is fundamentally different since it does not require domain knowledge and instead needs to verify adherence of the evidence to the rules provided in the prompt. 
A similar pattern appears when we look at calibration,
\Cref{fig:batch_vs_seq_ece} shows lower ECE in batch mode than BP mode for Animals, Political Ideology, and MediQ; 
the two modes have comparable calibration on Eleusis.

To explore this observed inconsistency between the batch and BP modes, next, we look at the information processing efficiency of both BP mode (\Cref{sec:bp_is_bayesian}) and batch mode (\Cref{sec:batch_isnt_bayesian}), the core study of this paper.

\subsection{LLMs often perform Bayesian updates in BP mode}\label{sec:bp_is_bayesian}

\begin{figure}[ht]
    \centering
    \includegraphics[width=0.9\linewidth]{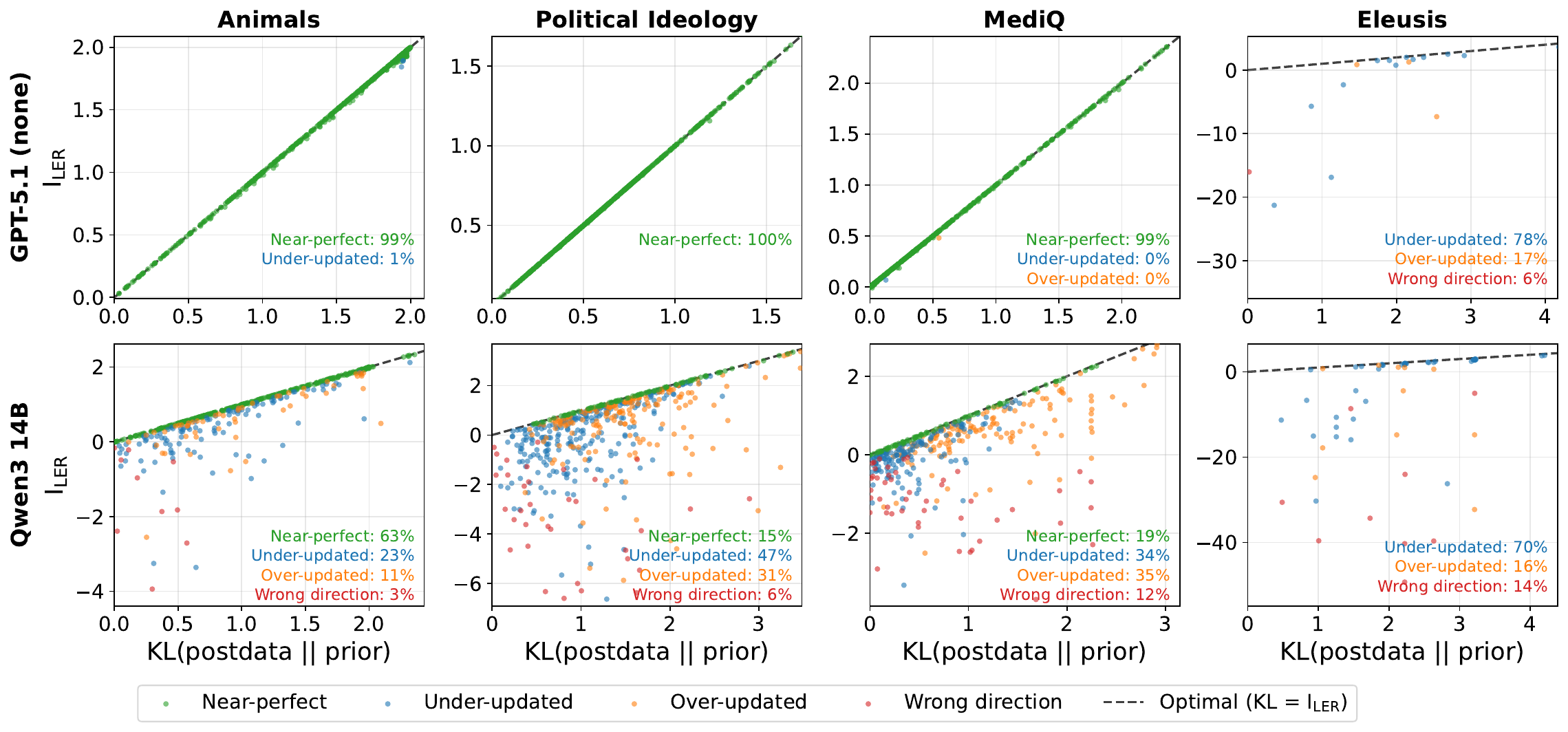}
    \caption{BP mode KL divergence vs.\ expected likelihood evidence ratio. Each point indicates an observation at the evidence step with $\geq 80\%$ valid observations (see text for details).
    }
    \label{fig:kl_vs_ler_seq}
\end{figure}

We explore information processing efficiency in BP mode by looking at the relationship between the expected likelihood evidence ratio $\ILER$ (\Cref{eq:Iler}) and the post-/pre-data KL divergence $\DKL(q||\pi)$ computed from the initial prior up to the max evidence step for each dataset; this is motivated by our definition of the information processing gap and its decomposition as detailed in \Cref{sec:decomposition}. \Cref{fig:kl_vs_ler_seq} shows the values of $\ILER$ and $\DKL$ for each question, for Qwen3 14B and GPT-5.1. Analogous plots for all models and evidence steps are shown in \Cref{fig:kl_vs_ler_steps_sequential_deepseekr10528qwen38b,fig:kl_vs_ler_steps_sequential_ministral38binstruct2512,fig:kl_vs_ler_steps_sequential_ministral38breasoning2512,fig:kl_vs_ler_steps_sequential_ministral314binstruct2512,fig:kl_vs_ler_steps_sequential_ministral314breasoning2512,fig:kl_vs_ler_steps_sequential_qwen38b,fig:kl_vs_ler_steps_sequential_qwen314b,fig:kl_vs_ler_steps_sequential_gpt4omini,fig:kl_vs_ler_steps_sequential_gpt51,fig:kl_vs_ler_steps_sequential_gpt51reasoningmedium}. In each plot, points are individual observations at the dataset-specific evidence step described earlier and the dashed line is where $\ILER=\DKL(q||\pi)$, implying optimal information processing.
Points near this line (specifically, with $\Delta(q) < 0.05$) are shown in green. 
Points colored red indicate wrong-direction updates (see \Cref{sec:decomposition}), 
where the new evidence has lower expected log-likelihood under the post-data distribution than under the prior. 
Points colored blue have under-updated, while points colored orange have over-updated, both in the right direction.
  
We note that, while the LLM's stated prior beliefs and likelihood model are included in-context, the LLM is not explicitly told to use Bayes' Theorem. Despite this, many points lie on, or close to, the dashed line, indicating optimal or near-optimal information processing (i.e., Bayesian updating). Indeed, the strongest model considered---GPT-5.1---is very close to optimal in all cases (with and without reasoning, see \Cref{fig:kl_vs_ler_steps_sequential_gpt51,fig:kl_vs_ler_steps_sequential_gpt51reasoningmedium}); DeepSeek-R1-0528-Qwen3-8B performs similarly (\Cref{fig:kl_vs_ler_steps_sequential_deepseekr10528qwen38b}). 

As we consider weaker models, and non-reasoning models, we see more deviations from the $\ILER=\DKL$ line. In \Cref{fig:kl_vs_ler_seq,fig:kl_vs_ler_steps_sequential_qwen314b}, we see that for Qwen3 14B, these updates are typically under-updated or over-updated relative to the likelihood, rather than moving in a direction inconsistent with the likelihood. Similar patterns are seen for both Ministral reasoning models (\Cref{fig:kl_vs_ler_steps_sequential_ministral38binstruct2512,fig:kl_vs_ler_steps_sequential_ministral314binstruct2512}). As we move towards weaker, non-reasoning models, we see many fewer optimal updates and an increasingly high number of wrong-direction updates, with the Ministral non-reasoning models being particularly prone to wrong-direction updates (\Cref{fig:kl_vs_ler_steps_sequential_ministral38binstruct2512,fig:kl_vs_ler_steps_sequential_ministral314binstruct2512,fig:kl_vs_ler_steps_sequential_qwen38b,fig:kl_vs_ler_steps_sequential_gpt4omini}). We hypothesize that this is a natural result of these models' weaker mathematical abilities, i.e.,  they are more likely to make mistakes when executing Bayes' Theorem, 
even when the prior and likelihood are provided in context. 

From this analysis, we can conclude that a reasonably strong LLM is able to perform optimal information processing provided they state their likelihoods explicitly and then update their beliefs one step at a time. 
However, does this extend to the batch setting, where we do not use the scaffolding of eliciting likelihoods and maintaining the intermediate belief distributions? We study this next.

\subsection{In batch mode, LLMs update their beliefs with implicit maps that are not consistent with their elicited likelihoods}

\label{sec:batch_isnt_bayesian}

\begin{figure}[ht]
    \centering
    \includegraphics[width=0.9\linewidth]{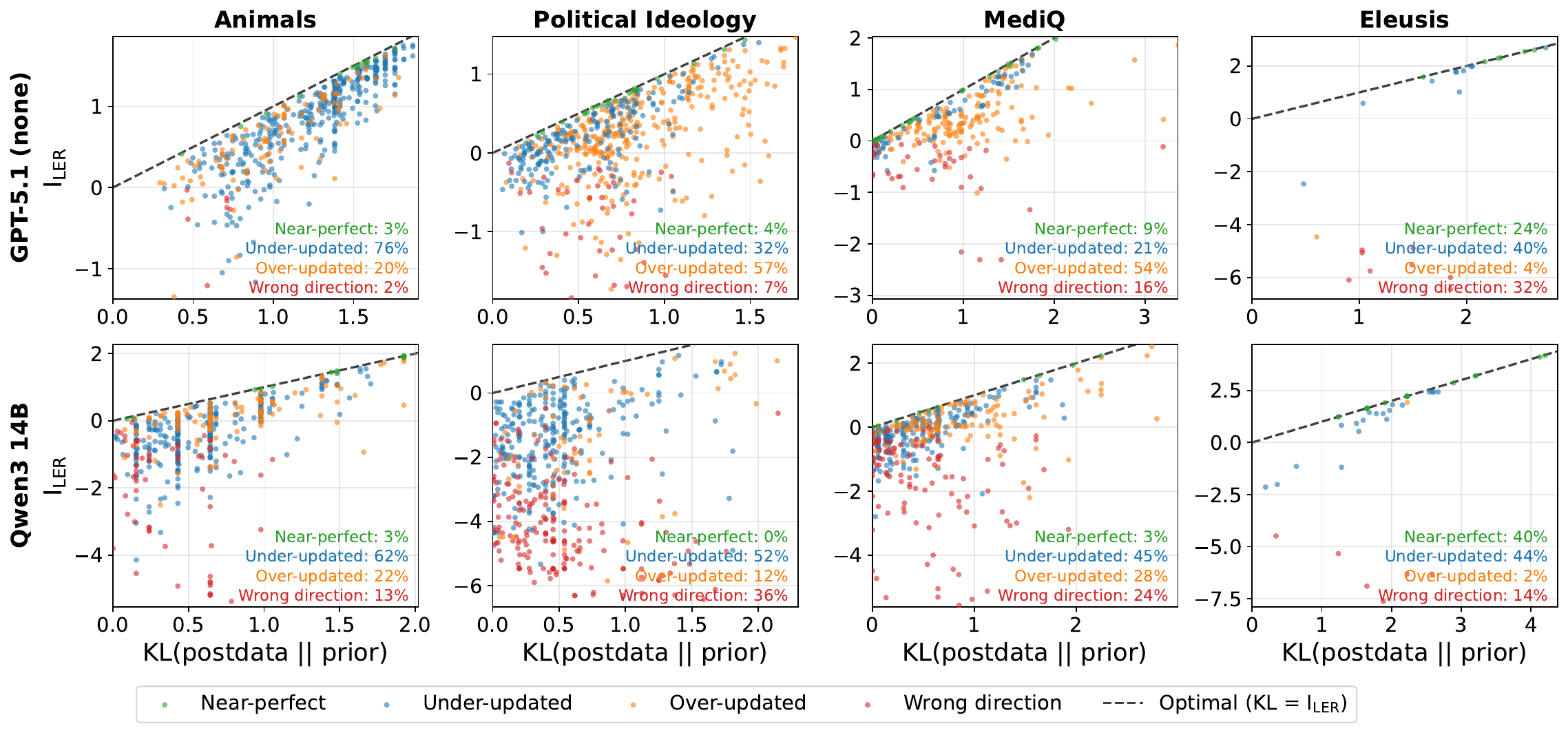}
    \caption{Batch mode KL divergence vs.\ expected likelihood evidence ratio. Each point indicates an observation at the evidence step with $\geq 80\%$ valid observations (see text for details).
    }
    \label{fig:kl_vs_ler_batch}
\end{figure}

\begin{figure}[ht]
\centering
\includegraphics[width=.99\linewidth]{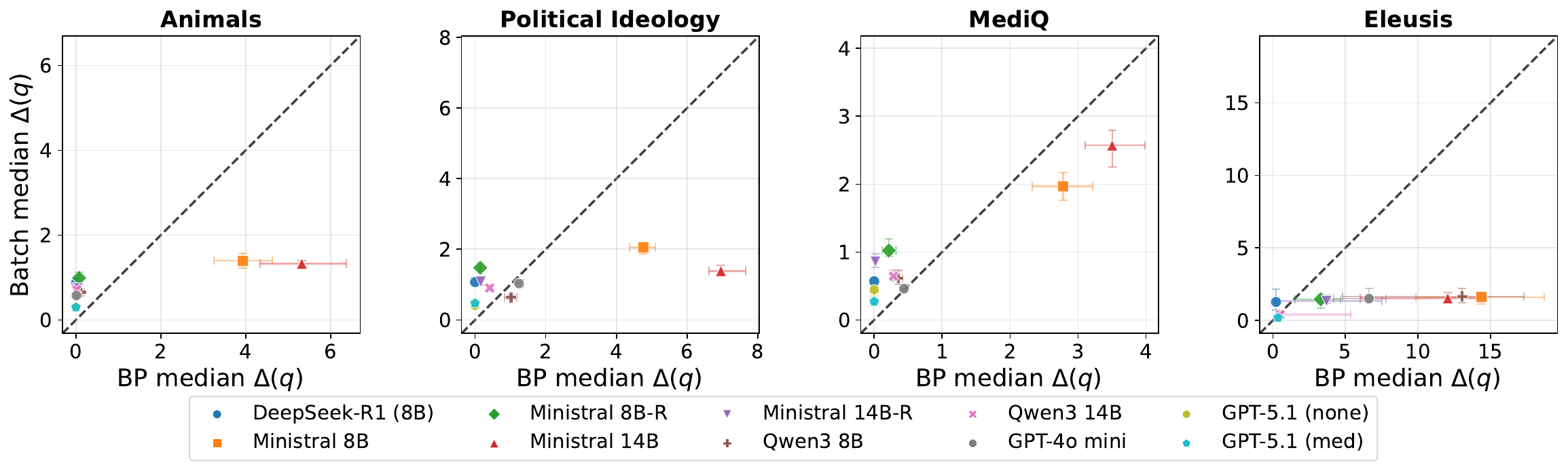}
\caption{Relationship between the information processing gap $\Delta(q)$ obtained in batch vs BP modes. }
\label{fig:batch_vs_seq_delta}
\end{figure}

We now repeat the analysis in \Cref{sec:bp_is_bayesian} using batch-mode belief updates. 
As before, the prompt does not explicitly direct the LLM to be Bayesian. 
Additionally, there is no encouragement to consider the evidence step-by-step.
In \Cref{fig:kl_vs_ler_batch}, we show the relationship between $\ILER$ and $\DKL$ for GPT-5.1 and Qwen3 14B (plots for all models, at all
evidence steps are shown in the Appendix in \Cref{fig:kl_vs_ler_steps_batch_gpt51,fig:kl_vs_ler_steps_batch_deepseekr10528qwen38b,fig:kl_vs_ler_steps_batch_gpt4omini,fig:kl_vs_ler_steps_batch_gpt51reasoningmedium,fig:kl_vs_ler_steps_batch_ministral314binstruct2512,fig:kl_vs_ler_steps_batch_ministral314breasoning2512,fig:kl_vs_ler_steps_batch_ministral38binstruct2512,fig:kl_vs_ler_steps_batch_ministral38breasoning2512,fig:kl_vs_ler_steps_batch_qwen314b,fig:kl_vs_ler_steps_batch_qwen38b}). Comparing with \Cref{fig:kl_vs_ler_seq,fig:kl_vs_ler_steps_sequential_deepseekr10528qwen38b,fig:kl_vs_ler_steps_sequential_ministral38binstruct2512,fig:kl_vs_ler_steps_sequential_ministral38breasoning2512,fig:kl_vs_ler_steps_sequential_ministral314binstruct2512,fig:kl_vs_ler_steps_sequential_ministral314breasoning2512,fig:kl_vs_ler_steps_sequential_qwen38b,fig:kl_vs_ler_steps_sequential_qwen314b,fig:kl_vs_ler_steps_sequential_gpt4omini,fig:kl_vs_ler_steps_sequential_gpt51,fig:kl_vs_ler_steps_sequential_gpt51reasoningmedium}, we see much higher rates of deviation from optimal information processing. Even GPT-5.1, which showed near-perfect Bayesian updates in BP mode (\Cref{fig:kl_vs_ler_seq,fig:kl_vs_ler_steps_sequential_gpt51,fig:kl_vs_ler_steps_sequential_gpt51reasoningmedium}), is frequently far from optimal in batch mode (\Cref{fig:kl_vs_ler_batch,fig:kl_vs_ler_steps_batch_gpt51,fig:kl_vs_ler_steps_batch_gpt51reasoningmedium}). In particular, we see a significant number of points that have moved in the wrong direction relative to the likelihood, or that have dramatically over-updated (indicated by a large $\DKL$ and a negative $\ILER$). This trend continues in most other models, with the weaker non-reasoning models showing very large deviations from optimality with many wrong-direction updates (\Cref{fig:kl_vs_ler_steps_batch_ministral38binstruct2512,fig:kl_vs_ler_steps_batch_ministral314binstruct2512}).

\Cref{fig:batch_vs_seq_delta} compares the median information gap in batch vs BP modes for each model and dataset; in most cases we have lower $\Delta(q)$ -- i.e., we are closer to Bayesian -- in BP mode. The two non-reasoning Ministral models are outliers here, with disproportionately large BP mode $\Delta(q)$ values. 
Per \Cref{sec:bp_is_bayesian}, these models do not appear to be able to consistently apply Bayes' Theorem even in BP mode.

It is not surprising that we see less optimal information processing in batch mode than in BP mode--- though the result is critical for understanding, deploying, and optimizing these systems.
In BP mode, due to its scaffolding, the LLM was explicitly updating its beliefs one piece of evidence at a time, and was provided numerical likelihoods that might trigger explicit mathematical application of the Bayes' Theorem by the model. 

The result of this section \textit{is} surprising in the context of \Cref{sec:task_performance}, where we saw that batch mode typically yields better task performance than BP mode. 
In other words, we are obtaining high-utility post-data beliefs, despite these beliefs not being consistent with the stated likelihoods. 
This suggests that the LLM is learning an \textit{implicit map} from evidence to post-data distribution, without explicitly formulating a likelihood model. 
In the next section, we explore the ways in which this implicit  mapping deviates from an explicit application of Bayes' Theorem.

\subsection{In batch mode, being Bayesian is not always better}

\begin{figure*}[h]
    \centering
    \includegraphics[width=0.9\linewidth]{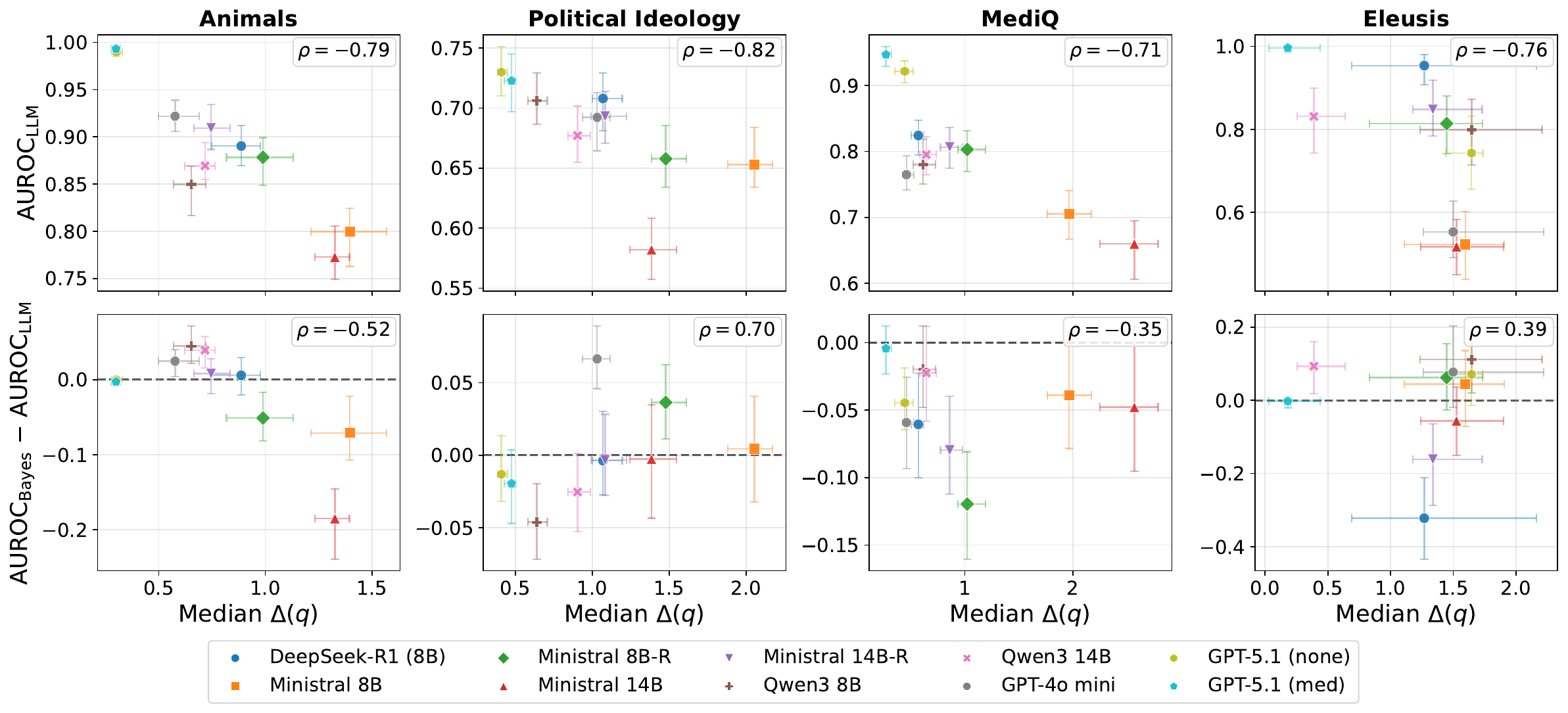}
    \caption{Exploring the relationship between task performance and information processing efficiency. Top: AUROC of LLM's batch-mode post-data distribution ($\text{AUROC}_\text{LLM}$), vs median information gap $\Delta(q)$, for each model. Bottom: Difference between $\text{AUROC}_{\text{Bayes}}$, the AUROC of the posterior obtained by explicitly applying Bayes' Theorem to the prior and likelihood elicited from the model, and $\text{AUROC}_\text{LLM}$, vs median information gap $\Delta(q)$; points below the dashed line indicate the LLM's batch post-data distribution outperforms optimal information processing, in terms of task performance. 
    }
    \label{fig:auroc_vs_delta_batch}
\end{figure*}

In \Cref{sec:batch_isnt_bayesian}, we found that the implicit map used in batch mode is often inconsistent with the elicited likelihoods. We now look at how this inconsistency relates to task performance.  In \Cref{fig:auroc_vs_delta_batch} (top), we show different model's AUROC and information processing gap $\Delta(q)$, evaluated on the maximum evidence step for each dataset. We first notice that AUROC tends to decrease with $\Delta(q)$. Superficially, this  suggests that we might get even better task performance if we forced the LLMs to be Bayesian. However, this correlation is subject to confounding: 
Is GPT-5.1 better because it has lower $\Delta(q)$, or are both the strong task performance and the small $\Delta(q)$ due to its size and training?

Since we know the model's prior $\pi$ and likelihood $\ell$, we can explicitly calculate the counterfactual task performance obtained using the Bayes posterior $p(\theta|X_{1:n}) \propto \pi(\theta) \prod_{i=1}^n \ell(X_i|\theta, X_{<i})$. In the bottom row of \Cref{fig:auroc_vs_delta_batch}, we show the difference in AUROC between the Bayes posterior $p$ ($\text{AUROC}_\text{Bayes}$) and the batch post-data distribution $q$ ($\text{AUROC}_\text{LLM}$), vs the median batch post-data gap $\Delta(q)$. 
We see that
there is not a clear causal relationship,
forcing $\Delta(q)=0$ does not always improve task performance (see \Cref{fig:auroc_vs_delta_bayes}). 
On MediQ, all models would perform worse if they were explicitly Bayesian. On Eleusis, the reasoning models (except Ministral 8B-R) would perform worse if they were explicitly Bayesian; however most other models would perform better. 
On Animals and Political Ideology, the explicit Bayes posterior yields no consistent performance improvement over the implicit mapping.

How do we explain these seeming contradictions? We hypothesize that this is a result of task structures that commonly appear in LLM training. We suspect that it is much more common to see training examples where a conclusion is drawn from multiple pieces of evidence, potentially as a result of complex reasoning, vs training examples where explicit priors and likelihoods are described. 
This hypothesis aligns with what we know about the four datasets. Specifying a statistical model for the MediQ dataset is challenging, since it requires significant knowledge of the rate at which different medical conditions lead to specific symptoms. However, LLMs have likely seen a large number of examples of medical diagnosis, making it plausible to learn a high-quality implicit mapping from data to beliefs. This combination of weak forward model and strong implicit inverse model would explain why forcing Bayesian updating degrades performance, in particular for very strong models.

Conversely, Eleusis is not a common game, and it is likely there isn't enough data for an LLM to learn a high-quality implicit mapping. However, the LLM \textit{is} given sufficient in-context information to construct a good likelihood model. In most cases, using this likelihood in an explicit Bayesian update outperforms the implicit mapping seen in batch mode. This is consistent with the task performance results shown in \Cref{fig:batch_vs_seq_auroc}, where Eleusis was the only dataset where batch mode did not yield better task performance than BP mode. An interesting observation is that the three models where the implicit batch-mode map \textit{does} outperform the explicit Bayes update on Eleusis are the three strong reasoning models; it is possible the additional reasoning capacity allows them to improve upon their initial statistical model, for example by reasoning about why certain moves were made. 

In the next section, we will explore structural properties of the post-data distributions, to better understand their behavior, and therefore that of LLMs,  when they deviate from optimal information processing.

\subsection{Updates that conflict with the elicited likelihood often align with an oracle likelihood} 

\begin{figure}[htb]
    \centering
    \includegraphics[width=0.9\linewidth]{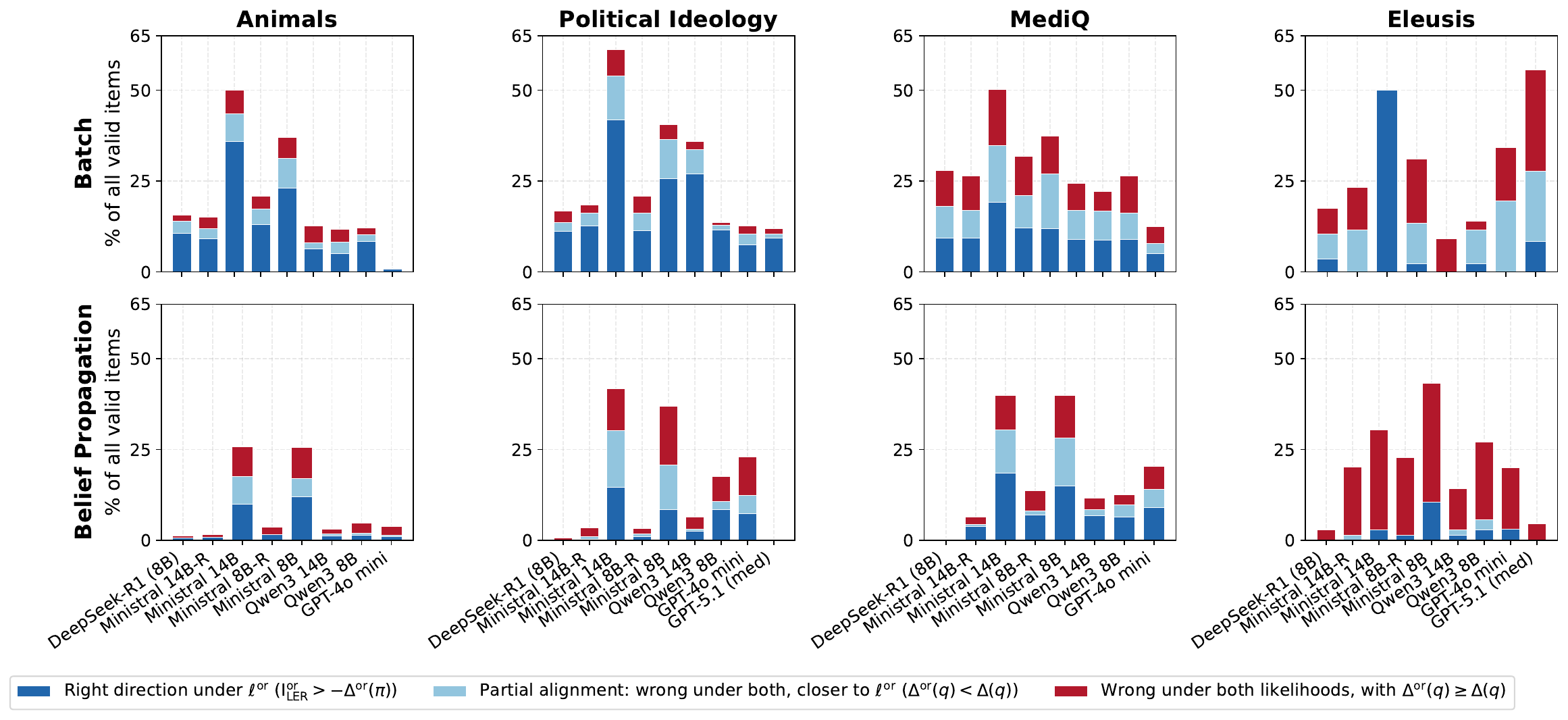}
    \caption{
    Breakdown of ``wrong-direction'' updates by alignment with oracle likelihood $\ell^\text{\tiny{or}}$. 
    }
    
    \label{fig:wrong_dir_updates}
\end{figure}
We now take a closer look at instances when the model's post-data updates move in the wrong direction relative to the model's likelihood (i.e., move mass toward beliefs not supported by the evidence). 
As we saw in \Cref{sec:decomposition}, this occurs when $\ILER \leq -\DKL(\pi || p)$. 
One hypothesis is that the elicited likelihood $\ell$ is misspecified, 
and that the model's update might be in fact more aligned with some ``ground truth'' likelihood of the data generating process. 
Here, we use GPT-5.1's likelihood as a proxy for this unknown true likelihood and dub it ``oracle'' $\ell^\text{\tiny{or}}$, 
since GPT-5.1 yielded 
the strongest task performance and smallest $\Delta(q)$ in both modes. 
We calculate 
$\Delta^\text{\tiny{or}}(q) = \DKL(q || p^\text{\tiny{or}})$,
where $p^\text{\tiny{or}}$ is the result of Bayes' Theorem applied to the model's own stated prior $\pi$ and GPT-5.1's likelihoods $\ell^\text{\tiny{or}}$. 

We can then analyze the ``wrong direction'' updates in terms of whether they are aligned with the oracle likelihood, i.e., whether $\ILER^\text{\tiny{or}} > -\DKL(\pi || p^\text{\tiny{or}})$, 
where $\ILER^\text{\tiny{or}}$ replaces the model's likelihood $\ell$ in \Cref{eq:Iler} with $\ell^\text{\tiny{or}}$. 
These updates are shown in dark blue in \Cref{fig:wrong_dir_updates}. 
For updates that are not supported by either the model's own likelihood or the oracle likelihood (i.e., where both $\ILER^\text{\tiny{or}} < -\DKL(\pi || p^\text{\tiny{or}})$ and $\ILER<-\DKL(\pi || p)$), we look at which posterior, $p^\text{\tiny{or}}$ or $p$, the model's post-data distribution $q$ is closer. 
If $\Delta(q)^\text{\tiny{or}} < \Delta(q)$ 
then the post-data belief is closer (in terms of KL) to the oracle posterior $p^\text{\tiny{or}}$ than to $p$; such points are shown in light blue.
If $\Delta(q)^\text{\tiny{or}} \ge \Delta(q)$, we mark them red.

We find that across all datasets but Eleusis, around half of the ``wrong direction'' updates in batch mode are actually moving in the direction implied by the oracle likelihood. 
Around 15\% more are moving counter to both likelihoods, but result in an update that is closer to $p^\text{\tiny{or}}$ than to $p$. 
This suggests that often, updates that conflict with the  model's own stated likelihood $\ell$ are more consistent with a better approximation to the ``ground truth'' likelihood.  
In  BP mode, wrong-direction updates are rare, suggesting that in this mode the model mostly follows its stated likelihood and disregards the implicit map knowledge the batch mode exploits. 
However, we again see that around half of the wrong-direction updates it makes are more aligned with the oracle posterior. 
In \Cref{app:oracle_auroc}, we additionally show that, for likelihoods that significantly deviate from the oracle, explicit Bayesian updating using the oracle likelihood leads to significant performance gains.

\subsection{Batch mode $\Delta(q)$ is informative of task performance in BP mode}

\begin{figure}[h]
    \centering
    \includegraphics[width=0.9\linewidth]{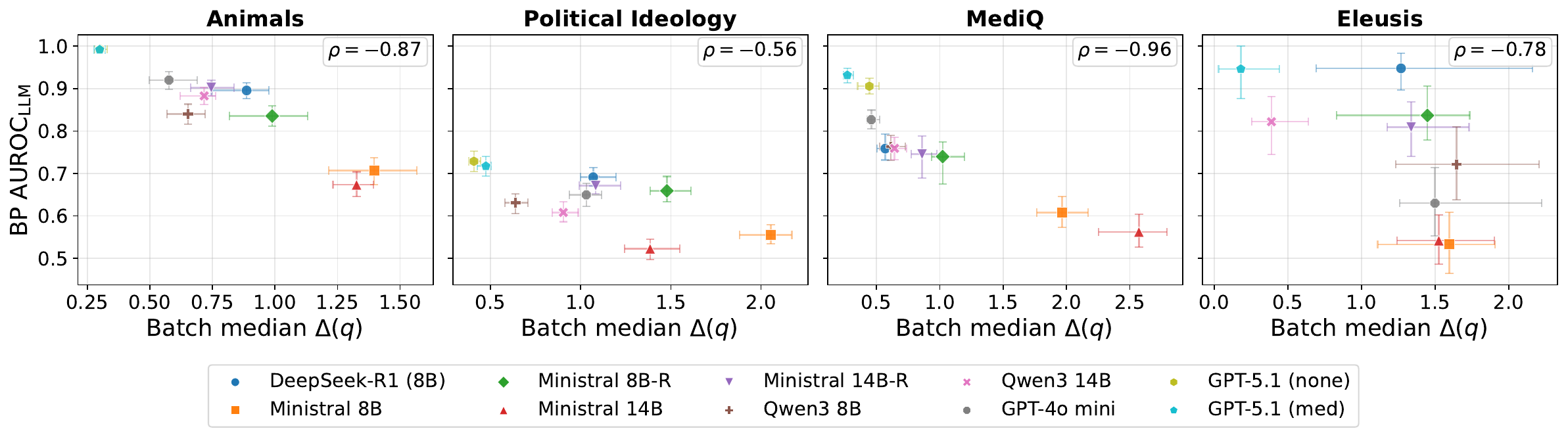}
    \caption{Relationship between BP-mode AUROC and batch-mode $\Delta(q)$.} 
    \label{fig:cross_mode_auroc_vs_delta}
\end{figure}

We know that  $\Delta(q)$ in BP mode is not going to give us much information about the corresponding task performance, since $\Delta(q)$ is often zero in this setting. Instead, in \Cref{fig:cross_mode_auroc_vs_delta}, we look at the relationship between BP mode AUROC and batch mode $\Delta(q)$. We find that batch mode $\Delta(q)$ is informative about BP mode task performance.

This relationship makes sense as we see small values of $\Delta(q)$ in batch mode when the model's likelihood is consistent with its implicit post-data distribution. 
This is most likely to occur when both the likelihoods are well-specified, and the implicit post-data belief mapping is of high quality,
suggesting that batch mode $\Delta(q)$ might be a good proxy for the quality of elicited likelihood estimates. 
This provides reassurance that BP performance is not being severely hampered by misspecified likelihoods.
We still face the confounding factor that stronger models will likely have both better likelihoods and more performant post-data distributions, 
however, we can be confident that the Bayesian behavior induced by BP mode is not compromising the model's performance.

\section{Related Work}
Several recent works have explored LLMs' abilities to reason about \textit{fully specified} statistical models \citep{jin2023cladder,aliakbar2025,schrader2024}, where a text description of a Bayesian network is given in-context. 
Our work goes beyond this setting, by formalizing the problem in an information-theoretic framework and requiring the LLM to construct its own statistical world model, we can probe the consistency and task performance of a model's beliefs. 

Closer to our work, \citet{qiu2026bayesian} argue that LLMs do not perform Bayesian updating by comparing LLM predictions to those of an optimal Bayesian model.  
\citet{pal2025incoherent} investigate LLMs' belief updates deviations from Bayesian updating in terms of Brier score. Our work provides a foundational framework to study the problem, supported by theory on optimal information processing, and illustrates that models being Bayesian or not is a more subtle question.

Other works assume (approximately) Bayesian updating to guide active information gathering \citep{choudhury2025,grand2025shootfirst,kobalczyk2025active}. 
This body of work is complementary to ours, by understanding when and how LLMs deviate from Bayesian updating, as formulated here, we can identify potential failure modes of these information gathering methods.

\section{Discussion}\label{sec:discussion}

Based on information theory and optimal information processing, we have shown that the information processing gap is a useful tool for quantifying how LLMs incorporate evidence when updating their beliefs. 
Our empirical results indicate that reasonably strong LLMs will often perform explicit Bayesian updates when presented with evidence and associated likelihoods in a sequential manner and instructed to update their beliefs after each piece of evidence. 
When presented with all evidence at once, we find that LLMs do not perform explicit Bayesian inference, but instead carry out a form of implicit inference to reach their post-data beliefs. 
Looking at the information gap of these implicitly derived post-data beliefs unearths surprising inconsistencies between these post-data beliefs and the likelihoods the LLM assigns to the evidence. 
Our findings have several implications for practitioners relying on LLM beliefs.

\paragraph{Implications.}

Despite the growing body of work calling for LLMs to operate in a Bayesian manner and specific works prompting LLMs to do so~\citep{nori2025sequential,shaikh2025creating}, our results indicate that you cannot offload good statistical practice to an LLM. In both our batch and BP modes we found that the LLM produces likelihoods that are not necessarily tied to the world, not necessarily tied to the post-data distribution, or both. Forcing a Bayesian update may in fact be harmful to task performance.

If all we care about is task performance from the posterior and have enough compute, in many cases the implicit mapping used in batch mode will perform well. However, we should be aware that the model may not be performing Bayesian updating consistent with its elicited likelihoods -- we shouldn't operate under the assumption that the stated likelihood has anything to do with the post-data distribution.

Alternatively, in many cases maintaining a belief state and updating it based on new evidence is desired. In this case practitioners should be careful. In many cases, LLMs do seem to perform optimal information processing -- i.e., Bayesian updating -- however, task performance often degrades because the likelihoods used do not seem to be a good model of the world. In this case, the batch mode information processing gap can be useful to diagnose potential issues and to gain confidence in the task performance of the approach.

\paragraph{Limitations.} $\Delta(q)$ is not comparable across evidence steps or datasets since the total amount of information may be different.
Additionally, $\Delta(q)$ is computed relative to likelihoods elicited through a specific prompt format and might be sensitive to that format.

In this first work analyzing LLMs as information processors, 
we studied the multiple-choice QA setting as a simplified version of real-world systems revealing surprising behaviors. Extending these ideas to more complex settings can reveal further capabilities and limitations of using LLMs for belief updating applications.

\newpage
\bibliographystyle{plainnat}
\bibliography{main}
\newpage

\appendix
\crefalias{section}{appendix}
\crefalias{subsection}{appendix}  

\renewcommand{\thefigure}{A\arabic{figure}}
\setcounter{figure}{0}
\renewcommand{\thetable}{A\arabic{table}}
\setcounter{table}{0}

\section{Prompts}
See \Cref{fig:prompt-templates-2} for the prompt templates used in the experiments and \Cref{fig:prompt-placeholder-values} for dataset-specific context provided to the LLMs.

\begin{figure}[ht]
    \centering
    \scriptsize
    \setlength{\fboxsep}{4pt}

    \newcommand{\stepinstr}{
        \texttt{First, provide a step-by-step explanation of your reasoning.
        Then, respond with the probability for each answer (A,\,B,\,C,\,D)
        as numbers between 0.0 and 1.0 summing to 1.0.}}

    \noindent\fbox{\parbox{\dimexpr\textwidth-2\fboxsep-2\fboxrule\relax}{
    \textbf{Likelihood prompt}
    \par\vspace{1pt}\hrule\vspace{2pt}
    \texttt{[SYSTEM]: \{dataset-specific system prompt\}}\\[2pt]
    \texttt{\{dataset-specific evidence wrapper\}: \{$\mathtt{X_1}$\}. \{$\mathtt{X_2}$\}. $\dots$ \{$\mathtt{X_n}$\}.}\\[2pt]
    \texttt{QUESTION: \{question\}}
    \texttt{~~A: \{$\mathtt{\theta_1}$\} ~B: \{$\mathtt{\theta_2}$\} ~C: \{$\mathtt{\theta_3}$\} ~D: \{$\mathtt{\theta_4}$\}}\\[2pt]
    \texttt{\{dataset-specific answer wrapper\}: \{$\mathtt{\theta_k}$\}}.\\[2pt]
    \texttt{\{dataset-specific new evidence wrapper\}: \{$\mathtt{X_{n+1}}$\}}\\[2pt]
    \texttt{\{dataset-specific task description\} 
    First, provide a step-by-step explanation of your reasoning. Then, output the probability as a number between 0.0 and 1.0.}
    }}

    \vspace{2pt}

    \noindent\fbox{\parbox{\dimexpr\textwidth-2\fboxsep-2\fboxrule\relax}{
    \textbf{Post-data belief prompt (Batch)}
    \par\vspace{1pt}\hrule\vspace{2pt}
    \texttt{\{dataset-specific evidence wrapper\}: \{$\mathtt{X_1}$\}. \{$\mathtt{X_2}$\}. $\dots$ \{$\mathtt{X_{n+1}}$\}.}\\[2pt]
    \texttt{QUESTION: \{question\}}
    \texttt{~~A: \{$\mathtt{\theta_1}$\} ~B: \{$\mathtt{\theta_2}$\} ~C: \{$\mathtt{\theta_3}$\} ~D: \{$\mathtt{\theta_4}$\}}\\[2pt]
    \stepinstr
    }}

    \vspace{2pt}

    \noindent\fbox{\parbox{\dimexpr\textwidth-2\fboxsep-2\fboxrule\relax}{
    \textbf{Post-data belief prompt (Belief propagation)}
    \par\vspace{1pt}\hrule\vspace{2pt}
    \texttt{QUESTION: \{question\}}
    \texttt{~~A: \{$\mathtt{\theta_1}$\} ~B: \{$\mathtt{\theta_2}$\} ~C: \{$\mathtt{\theta_3}$\} ~D: \{$\mathtt{\theta_4}$\}}\\[2pt]
    \texttt{\{dataset-specific prior wrapper\}:}
    \texttt{~~A: \{$\pi_1$\} ~B: \{$\pi_2$\} ~C: \{$\pi_3$\} ~D: \{$\pi_4$\}}\\[2pt]
    \texttt{You now observe the new evidence: \{$X_{n+1}$\}}\\[2pt]
    \noindent
        \texttt{\{dataset-specific likelihood wrapper:
        A: \{$\ell_1$\} ~\ldots~ D: \{$\ell_4$\}}
    \\[2pt]
    \stepinstr
    }}

\caption{Prompt templates for all datasets.}
\label{fig:prompt-templates-2}
\end{figure}

\begin{figure}[p]
    \centering
    \scriptsize
    \setlength{\fboxsep}{4pt}

\noindent\fbox{\parbox{\dimexpr\textwidth-2\fboxsep-2\fboxrule\relax}{
\textbf{Dataset-specific placeholder values}
\par\vspace{1pt}\hrule\vspace{2pt}
\textbf{Animals}\\[1pt]
\begin{tabular}{@{} l @{\hspace{6pt}} p{0.8\linewidth} @{}}
System prompt        & \texttt{You are a helpful assistant who is an expert on identifying
                       animals from their attributes. You will be presented with
                       various attributes that an animal possesses or not, and your
                       task will be to identify the animal as best you can.} \\
Evidence wrapper     & \texttt{ANIMAL ATTRIBUTES} \\
Answer wrapper       & \texttt{The correct answer to the question above is} \\
New evidence wrapper & \texttt{Consider the following piece of new information} \\
Task description     & \texttt{Given the correct answer and the provided information, your task is to
                       determine the probability of observing the new information.} \\
Prior wrapper        & \texttt{Your current beliefs about the animals are} \\
Likelihood wrapper   & \texttt{The likelihoods of this attribute being observed for each animal are} \\
\end{tabular}
\par\vspace{4pt}\hrule\vspace{2pt}
\textbf{Political Ideology}\\[1pt]
\begin{tabular}{@{} l @{\hspace{6pt}} p{0.8\linewidth} @{}}
System prompt        & \texttt{You are a social science reasoning assistant. You will receive
                       a set of survey responses from an individual and a multiple-choice
                       question about their demographic background. Each survey response
                       is formatted as <|Q|>: <question text> <|A|>: <answer text>.} \\
Evidence wrapper     & \texttt{SURVEY RESPONSES} \\
Answer wrapper       & \texttt{The actual demographic for this respondent is} \\
New evidence wrapper & \texttt{Consider the following survey response} \\
Task description     & \texttt{Given the respondent's actual demographic and the previous survey responses,
                       your task is to determine the probability of observing this survey response.} \\
Prior wrapper        & \texttt{Your current beliefs about the respondent's demographic are} \\
Likelihood wrapper   & \texttt{The likelihoods of this survey response being observed for each demographic are} \\
\end{tabular}
\par\vspace{4pt}\hrule\vspace{2pt}
\textbf{MediQ}\\[1pt]
\begin{tabular}{@{} l @{\hspace{6pt}} p{0.8\linewidth} @{}}
System prompt        & \texttt{You are a medical doctor trying to reason through a real-life clinical case.
                       Based on your understanding of basic and clinical science, medical knowledge,
                       and mechanisms underlying health, disease, patient care, and modes of therapy,
                       respond according to the task specified by the user. Base your response on
                       the current and standard practices referenced in medical guidelines.} \\
Evidence wrapper     & \texttt{A patient comes into the clinic presenting with the following symptoms: PATIENT INFORMATION} \\
Answer wrapper       & \texttt{The correct answer to the question above is} \\
New evidence wrapper & \texttt{Consider the following piece of new information} \\
Task description     & \texttt{Given the correct answer and the provided information, your task is to
                       determine the probability of observing the new information.} \\
Prior wrapper        & \texttt{Your current beliefs about the diagnosis are} \\
Likelihood wrapper   & \texttt{The likelihoods of this symptom being observed for each diagnosis are} \\
\end{tabular}
\par\vspace{4pt}\hrule\vspace{2pt}
\textbf{Eleusis}\\[1pt]
\begin{tabular}{@{} l @{\hspace{6pt}} p{0.8\linewidth} @{}}
System prompt        & \texttt{You are a helpful assistant who is an expert in logical reasoning
                       and pattern recognition. You will be presented with sequences of
                       card plays from a card game where cards are either accepted or
                       rejected according to a secret rule. Your task is to determine
                       which rule governs which cards are accepted.} \\
Evidence wrapper     & \texttt{CONTEXT} \\
Answer wrapper       & \texttt{The correct answer to the question above is} \\
New evidence wrapper & \texttt{Consider the following piece of new information} \\
Task description     & \texttt{Given the correct answer and the provided information, your task is to
                       determine the probability of observing the new information.} \\
Prior wrapper        & \texttt{Your current beliefs about the rule are} \\
Likelihood wrapper   & \texttt{The likelihoods of this card play being observed for each rule are} \\
\end{tabular}
}}

    \caption{Dataset-specific context that is provided to the LLMs in the prompt templates from \Cref{fig:prompt-templates-2}.}
    \label{fig:prompt-placeholder-values}
\end{figure}

\section{Additional Experimental Details}\label{app:experimental_details}

\subsection{Datasets}\label{app:datasets}
\begin{figure}
    \centering
    \includegraphics[width=0.99\linewidth]{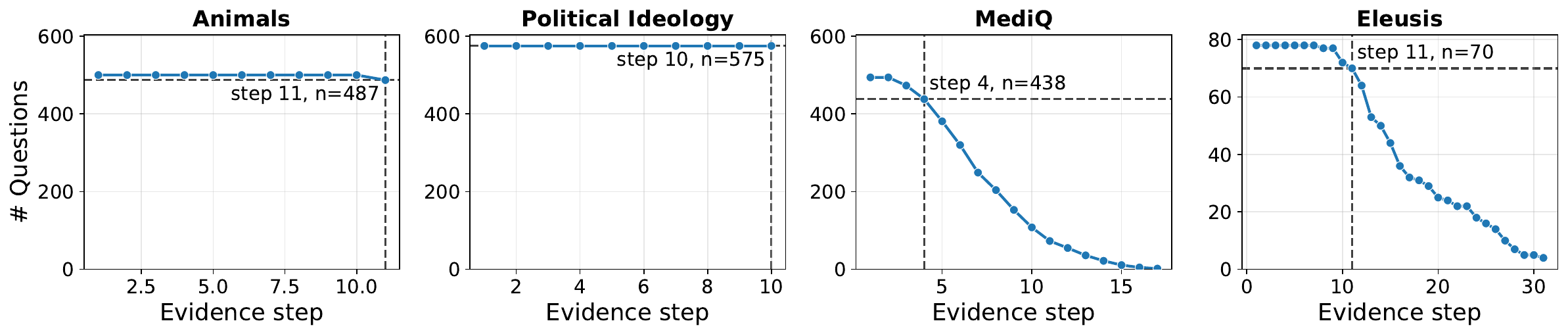}
    \caption{The number of questions for each dataset with observed evidence at each evidence step. The vertical dashed lines indicate the maximal evidence step selected using the filter described in \Cref{sec:results}.}
    \label{fig:n_questions_trajectories}
\end{figure}

Below we provide further details about the four datasets used in this paper. A summary table is provided in \Cref{tab:dataset_statistics}. \Cref{fig:n_questions_trajectories} shows the number of questions at each evidence step for each dataset and the maximal evidence steps which $\geq 80\%$ of the observations which were used to generate the plots in the main text.

\begin{table}[h]
\centering
\caption{Dataset Statistics}
\label{tab:dataset_statistics}
\begin{tabular}{lllp{3cm}p{3cm}}
\toprule
\textbf{Dataset} & \textbf{Datapoints} & \textbf{Options} & \textbf{Evidence Steps} & \textbf{80\% Evidence Step} \\
\midrule
Animals with Attributes & 500 & 4 & 11 & 11 \\
OpinionQA & 575 & 5 & 10 & 10\\
MediQ & 494$^*$ & 4--10 & variable & 4\\
Eleusis & 78 & 5 & 4--31 & 11 \\
\bottomrule
\end{tabular}
\vspace{0.2cm}
\end{table}

The following figure indicates the number of observations with evidence at the specified evidence step. The Animals and Political Ideology datasets have all observations contain the same amount of evidence, but MediQ and Eleusis drop off quickly.

\begin{landscape}
\tiny
\begin{longtable}{llllllllllll}
\caption{Animals dataset. Each row represents a species and its associated attribute values.}\label{tab:awa-full} \\
\toprule
\textbf{Species} & \textbf{Color} & \textbf{Food} & \textbf{Habitat} & \textbf{Locomotion} & \textbf{Reproduces} & \textbf{Active} & \textbf{Exterior} & \textbf{Pattern} & \textbf{Social structure} & \textbf{Size} & \textbf{Extremities} \\
\midrule
\endfirsthead
\toprule
\textbf{Species} & \textbf{Color} & \textbf{Food} & \textbf{Habitat} & \textbf{Locomotion} & \textbf{Reproduces} & \textbf{Active} & \textbf{Exterior} & \textbf{Pattern} & \textbf{Social structure} & \textbf{Size} & \textbf{Extremities} \\
\midrule
\endhead
\midrule
\multicolumn{12}{r}{\textit{Continued on next page}} \\
\endfoot
\bottomrule
\endlastfoot
Lion & brown & vertebrate & grassland & terrestrial & live birth & nocturnal & fur & solid & pack & large & paws \\
Tiger & orange & vertebrate & forest & terrestrial & live birth & nocturnal & fur & striped & solitary & large & paws \\
Cheetah & yellow & vertebrate & grassland & terrestrial & live birth & diurnal & fur & spotted & solitary & large & legs \\
Cougar & brown & vertebrate & forest & terrestrial & live birth & crepuscular & fur & solid & solitary & large & paws \\
Canada Lynx & gray & vertebrate & forest & terrestrial & live birth & nocturnal & fur & marked & -- & medium & paws \\
Polar Bear & white & vertebrate & arctic & amphibious & live birth & cathemeral & fur & solid & solitary & huge & paws \\
Grizzly Bear & brown & vertebrate & forest & terrestrial & live birth & diurnal & fur & mottled & solitary & large & paws \\
American Black Bear & black & plant & forest & terrestrial & live birth & crepuscular & fur & solid & solitary & large & paws \\
Sun Bear & black & invertebrate & forest & terrestrial & live birth & diurnal & fur & marked & solitary & medium & paws \\
Sloth Bear & black & invertebrate & forest & terrestrial & live birth & nocturnal & fur & marked & solitary & large & paws \\
Gray Wolf & gray & vertebrate & forest & terrestrial & live birth & crepuscular & fur & mottled & pack & large & paws \\
Red Fox & red & vertebrate & forest & terrestrial & live birth & nocturnal & fur & marked & solitary & small & paws \\
Coyote & gray & vertebrate & grassland & terrestrial & live birth & nocturnal & fur & mottled & pack & medium & paws \\
Arctic Fox & white & vertebrate & arctic & terrestrial & live birth & cathemeral & fur & solid & pair & small & paws \\
Fennec Fox & yellow & invertebrate & desert & terrestrial & live birth & nocturnal & fur & solid & pack & small & paws \\
Western Gorilla & black & fruit & forest & terrestrial & live birth & diurnal & fur & solid & troop & large & hands \\
Common Chimpanzee & black & fruit & forest & terrestrial & live birth & diurnal & fur & solid & troop & large & hands \\
Bornean Orangutan & orange & fruit & forest & terrestrial & live birth & diurnal & fur & solid & solitary & large & hands \\
Olive Baboon & gray & plant & grassland & terrestrial & live birth & diurnal & fur & mottled & troop & medium & hands \\
Japanese Macaque & brown & plant & forest & terrestrial & live birth & diurnal & fur & solid & troop & medium & hands \\
Blue Whale & blue & invertebrate & ocean & aquatic & live birth & cathemeral & skin & mottled & solitary & huge & flippers \\
Sperm Whale & gray & invertebrate & ocean & aquatic & live birth & cathemeral & skin & solid & pod & huge & flippers \\
Common Bottlenose Dolphin & gray & vertebrate & ocean & aquatic & live birth & cathemeral & skin & countershading & pod & large & flippers \\
Beluga Whale & white & vertebrate & arctic & aquatic & live birth & cathemeral & skin & solid & pod & large & flippers \\
Narwhal & gray & vertebrate & arctic & aquatic & live birth & cathemeral & skin & mottled & pod & large & flippers \\
Walrus & brown & invertebrate & arctic & amphibious & live birth & cathemeral & skin & solid & colony & huge & flippers \\
California Sea Lion & brown & vertebrate & ocean & amphibious & live birth & cathemeral & fur & solid & colony & large & flippers \\
Leopard Seal & gray & vertebrate & ocean & amphibious & live birth & diurnal & fur & spotted & solitary & large & flippers \\
Northern Elephant Seal & brown & invertebrate & ocean & amphibious & live birth & cathemeral & fur & solid & colony & huge & flippers \\
Harbor Seal & gray & vertebrate & ocean & amphibious & live birth & cathemeral & fur & spotted & colony & medium & flippers \\
King Cobra & brown & vertebrate & forest & slithering & lays eggs & diurnal & scales & striped & solitary & large & none \\
Green Anaconda & green & vertebrate & freshwater & amphibious & live birth & nocturnal & scales & spotted & solitary & huge & none \\
Boa Constrictor & brown & vertebrate & forest & slithering & live birth & nocturnal & scales & geometric & solitary & large & none \\
Western Diamondback Rattlesnake & brown & vertebrate & desert & slithering & live birth & nocturnal & scales & geometric & solitary & medium & none \\
Burmese Python & brown & vertebrate & forest & slithering & lays eggs & nocturnal & scales & mottled & solitary & huge & none \\
Great White Shark & gray & vertebrate & ocean & aquatic & live birth & diurnal & scales & countershading & solitary & huge & fins \\
Whale Shark & gray & invertebrate & ocean & aquatic & live birth & cathemeral & skin & spotted & solitary & huge & fins \\
Great Hammerhead Shark & gray & vertebrate & ocean & aquatic & live birth & cathemeral & scales & countershading & solitary & large & fins \\
Tiger Shark & gray & vertebrate & ocean & aquatic & live birth & nocturnal & scales & striped & solitary & large & fins \\
Bull Shark & gray & vertebrate & ocean & aquatic & live birth & cathemeral & scales & countershading & solitary & large & fins \\
Black Widow Spider & black & invertebrate & forest & terrestrial & lays eggs & nocturnal & exoskeleton & marked & solitary & small & legs \\
Emperor Scorpion & black & invertebrate & forest & terrestrial & live birth & nocturnal & exoskeleton & solid & colony & small & legs \\
Brown Recluse Spider & brown & invertebrate & forest & terrestrial & lays eggs & nocturnal & exoskeleton & marked & solitary & small & legs \\
Goliath Birdeater & brown & invertebrate & forest & terrestrial & lays eggs & nocturnal & exoskeleton & solid & solitary & large & legs \\
European Garden Spider & brown & invertebrate & forest & terrestrial & lays eggs & nocturnal & exoskeleton & geometric & solitary & small & legs \\
Mute Swan & white & plant & freshwater & amphibious & lays eggs & diurnal & feathers & solid & pair & large & wings \\
American Crow & black & invertebrate & forest & aerial & lays eggs & diurnal & feathers & solid & flock & medium & wings \\
Golden Eagle & brown & vertebrate & forest & aerial & lays eggs & diurnal & feathers & solid & pair & large & wings \\
Greater Flamingo & pink & invertebrate & freshwater & aerial & lays eggs & diurnal & feathers & solid & colony & large & wings \\
Emu & brown & plant & grassland & terrestrial & lays eggs & diurnal & feathers & solid & flock & large & legs \\
\end{longtable}
\end{landscape}
\paragraph{Animals} We consider a species identification task where an LLM must report its beliefs about what an unknown species is given only some attributes of it. The LLM must utilize its intrinsic knowledge about animals to inform its beliefs. We structure the dataset similarly to the \textit{Animals with Attributes} dataset which has been previously studied in the machine learning literature~\citep{osherson1991default,lampert2013attribute,kemp2006learning}. We prompted Gemini 2.5 Pro (on Oct. 6, 2025) to generate 50 species names (such that multiple species share the same phylogenetic group) and their values for 11 prespecified attributes – e.g., color, habitat, etc. The full dataset is shown in \Cref{tab:awa-full}. We generate a set of 500 questions each with four answer choices by selecting a target species and then selecting three other options at random with at least one coming from the same family – e.g. bear. This construction mitigates the computational scaling of the Zellner computations with the number of options while also making the questions challenging. We compute an approximate ground-truth posterior for each question by finding all species in the answer options that are consistent with the evidence and normalizing to be a probability distribution over that subset of species.

\paragraph{Political Ideology}

We construct a task identifying the political ideology of individuals who responded to multiple questionnaires by the Pew American Trends Panel. Our task is based on one proposed to evaluate training LLMs to optimally gather information~\citep{santurkar2023opinions}. The responses and ideology information we used were taken from the OpinionQA dataset \citep{santurkar2023opinions} which was originally developed to study whose opinions were being encoded into LLMs.

We build our dataset by identifying ten waves of the questions each with a different theme to provide different information about individuals. We chose ten waves to capture a broad set of question themes without too much overlap and to maintain enough overlapping respondents between the waves to perform a meaningful analysis. We retained the responses of all respondents who participated in nine out of the ten selected waves, yielding 575 respondents each with 10 evidence steps. Within each wave we used a Naive Bayes classifier to determine how predictive each question is of the respondents' political ideologies. We found that the most predictive questions within each wave topped out at $\sim0.4$ accuracy. We chose the most predictive question from each wave as the representative evidence for that wave. 
We then convert each data point into a multiple-choice question answering task by making the five political ideologies the answer options and each question and response from each wave the evidence for the question ``Based on the responses, what is the individual's political ideology?".

\paragraph{MediQ}
MediQ is an interactive clinical reasoning benchmark designed to evaluate an LLM's ability to ask information-seeking questions~\citep{li2024mediq}. While we do not evaluate models on an interactive question-asking setting, we use MediQ's datasets and processing pipeline. In particular, each question $(X, Y)$, where $X$ is a paragraph describing the patient's background and symptoms, and $Y$ is a ground truth diagnosis, is partitioned into a sequence of atomic evidence statements $X = \{X_0, X_1, ..., X_n\}$. While MediQ uses this evidence partitioning to instantiate a user simulator, we leverage this partitioning to probe sequential information processing capabilities. We evaluate on a randomly sampled subset of 500 datapoints from the MedQA-dev split of MediQ~\citep{jin2021disease}, a popular medical benchmark involving diagnostic questions drawn from medical board exams. We focus on the MedQA subset to evaluate general medical reasoning capabilities rather than domain-specific subsets.

This setting aims to emulate a doctor sequentially updating their beliefs about a patient's disease status in light of new evidence. We caution that this setting is not intended to faithfully model how human clinicians make diagnoses: real physicians ask questions, consult external resources and colleagues, and rely on heuristics. However, this setting provides a controlled environment for evaluating sequential evidence integration in natural language over a complex hypothesis space in a task of real-world significance. Importantly, eventual AI medical systems should update their beliefs in a coherent, Bayesian manner, rather than inheriting human diagnostic biases.

\paragraph{Eleusis}

The Eleusis game has recently been used to study how LLMs can act as scientists~\citep{louapre2026_can_llms_play_the_game_of_science}. 
The simplified variant of the game used in that work uses a standard 52-card deck and starts by selecting a rule from a fixed set of 26 rules which is hidden to the player – e.g. cards with Diamond suit – that they must infer from evidence. Players select cards to play to acquire information about the rule and they are given feedback whether the played card adheres to the rule (put on the ``main line") or does not (moved to a ``side line"). The aforementioned study evaluated the performance of a number of frontier LLMs on this variant of the game. Rather than have LLMs play the game, we utilize the traces from the Eleusis LLM Benchmark data~\citep{louapre2026_can_llms_play_the_game_of_science} and evaluate how efficiently the LLMs process the information.
We analyze the 78 game traces in the benchmark that were generated by GPT-5.2 with high reasoning. The traces contained 4--31 evidence steps. Note that we do not evaluate GPT-5.2's information processing capabilities.
Since the card-playing strategy of the original player determines which evidence is available, the information content of each split varies depending on how effectively the original player acquired evidence.

We construct a multiple-choice question from each game trace. We take the correct rule and sample four distractor rules to form the answer options. The evidence steps are the played cards and the associated lines indicating adherence to the hidden rule. We also supply the LLMs with the rules of the game so that they know how to interpret the information.

\subsection{Metrics}\label{app:metrics}

The information gap $\Delta(q)$ looks at whether the post-data distribution $q$ is consistent with the information implied by the prior and likelihood. However, a low value of $\Delta(q)$ does not mean that $q$ is an objectively ``good'' or useful distribution, since the prior or likelihood may be misspecified. We therefore consider two additional metrics that look at the utility of 
$q$ in realistic tasks. First, we use a multiple-choice
extension of AUROC. Interpreting AUROC as the Mann-Whitney statistic, we
extend it to the multiple-choice setting as the probability that a randomly
selected correct option (pooled across the dataset) is assigned higher
probability than a randomly selected incorrect option. This captures how
well a model's beliefs discriminate between correct and incorrect
hypotheses. Second, we use expected calibration error
\citep[ECE,][]{guo2017calibration}, which measures whether the model's
beliefs can be meaningfully interpreted as probabilities in a frequentist
sense. 

\subsection{Computation}\label{app:compute}
Closed-source models were accessed using API calls. Open-source models were run on A100 GPUs. Runtimes for the experiments varied from on the order of hours to days, depending on the dataset, mode (batch or belief propagation), and model being evaluated.

\subsection{Licensing}\label{licensing}
The Political Ideology dataset is a subset of the OpinionQA dataset \citep{santurkar2023opinions}, released under an MIT license. The Eleusis dataset is a subset of the dataset introduced in \citet{louapre2026_can_llms_play_the_game_of_science}, released under an MIT license. The MediQ dataset \citep{li2024mediq} is released under a CC-by-4.0 license.

The Ministral and Qwen3 models used in this paper are released under an Apache-2.0 license. DeepSeek-R1-0528-Qwen3-8B is released under an MIT license.

\FloatBarrier
\section{Additional Experimental Results}\label{sec:addl_experiments}

\subsection{Task performance vs.\ evidence step}

\begin{figure*}[htb]
    \centering
    \includegraphics[width=0.99\linewidth]{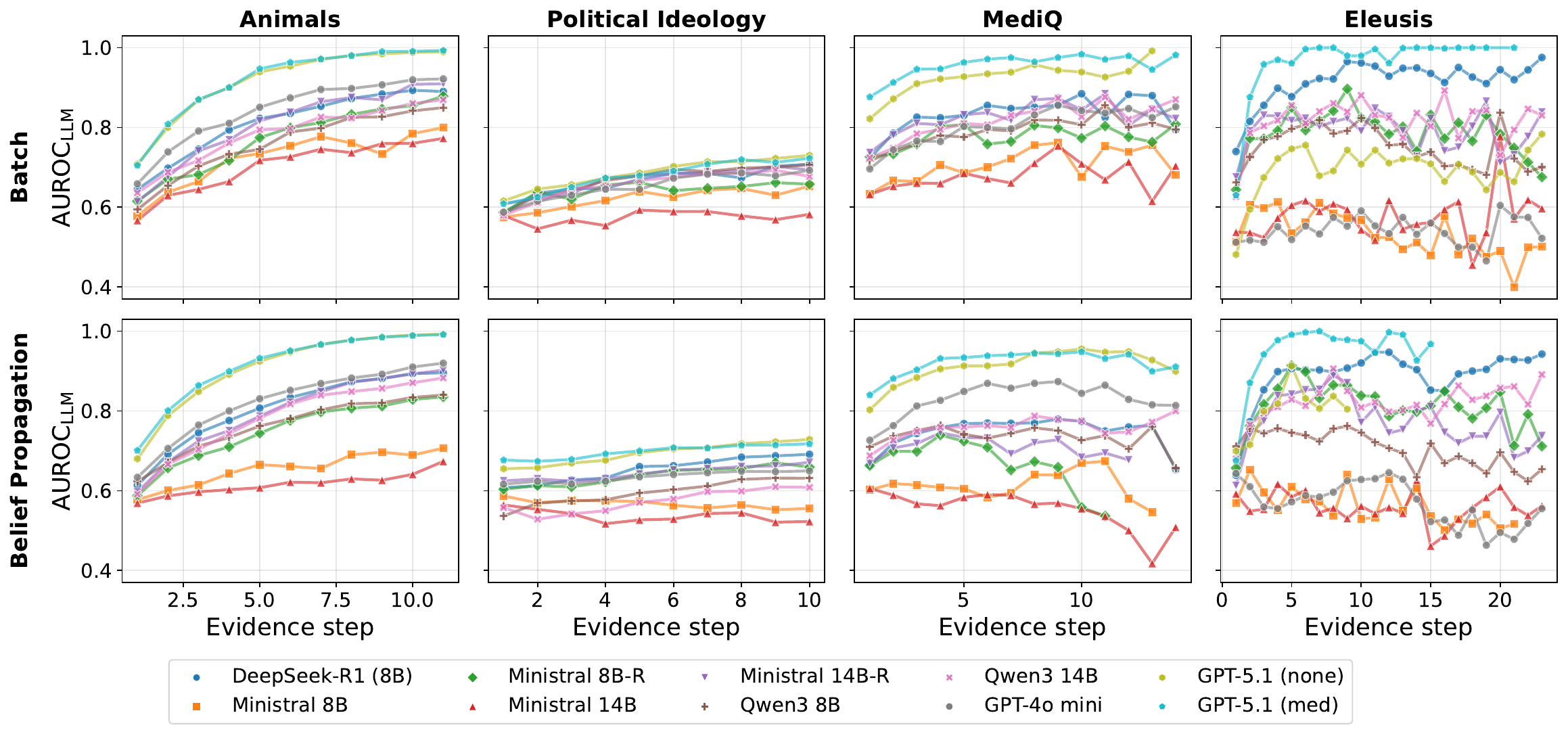}
    \caption{AUROC vs.\ evidence step using the LLM elicited post-data distribution.}
    \label{fig:auroc_evidence_step}
\end{figure*}

\begin{figure*}[htb]
    \centering
    \includegraphics[width=0.99\linewidth]{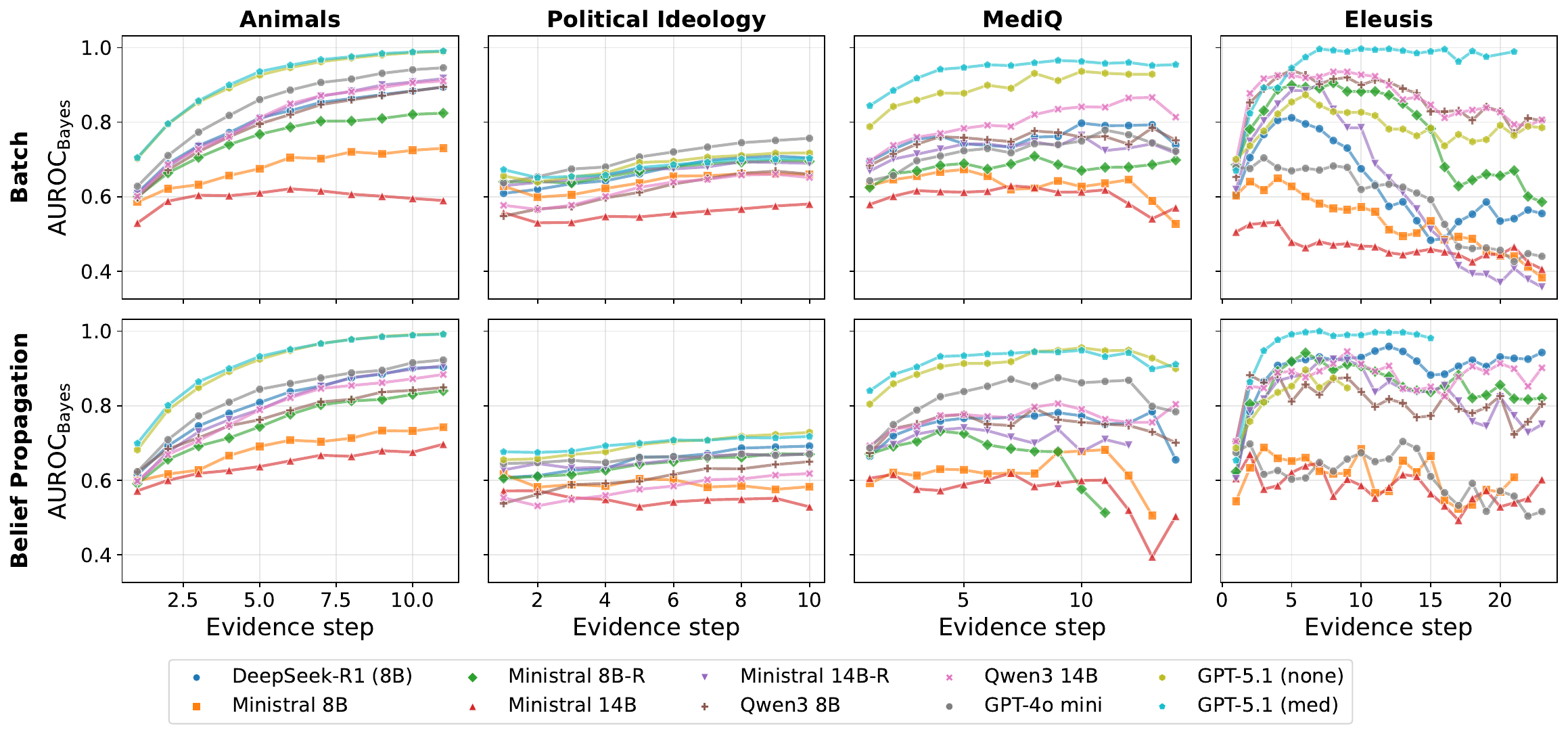}
    \caption{AUROC vs.\ evidence step using the Bayes posterior with LLM elicited likelihood and prior post-data distribution.}
    \label{fig:auroc_bayes_evidence_step}
\end{figure*}

In \Cref{fig:auroc_evidence_step} we show how AUROC varies across evidence step in both batch and BP modes, as discussed in \Cref{sec:task_performance}. In \Cref{fig:auroc_bayes_evidence_step}, we repeat this analysis using the explicit Bayes posterior $p$ where we manually compute Bayes' Theorem using the LLM's elicited prior and likelihood. We find similar performance over evidence steps from both approaches.

\begin{figure*}[htb]
    \centering
    \includegraphics[width=0.99\linewidth]{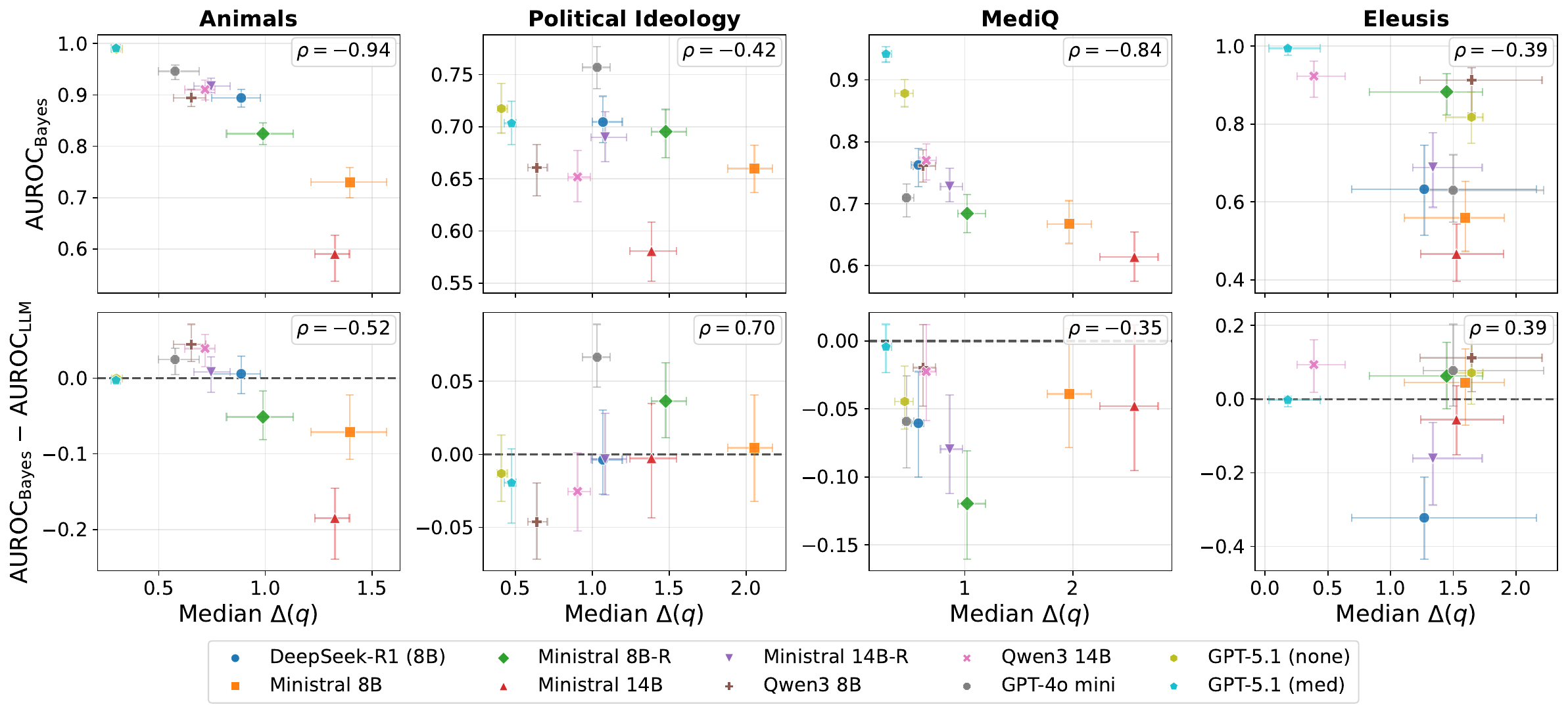}
    \caption{AUROC vs.\ $\Delta$ using the Bayes posterior with LLM elicited likelihood and the previous post-data distribution as the prior. This is equivalent to forcing $\Delta(q) = 0$ using the LLM's internal prior and likelihood.}
    \label{fig:auroc_vs_delta_bayes}
\end{figure*}

\subsection{Comparing model calibration in batch vs.\ belief propagation modes}
\label{app:calibration_batch_vs_seq}
\begin{figure}[htb]
    \centering
    \includegraphics[width=0.99\linewidth]{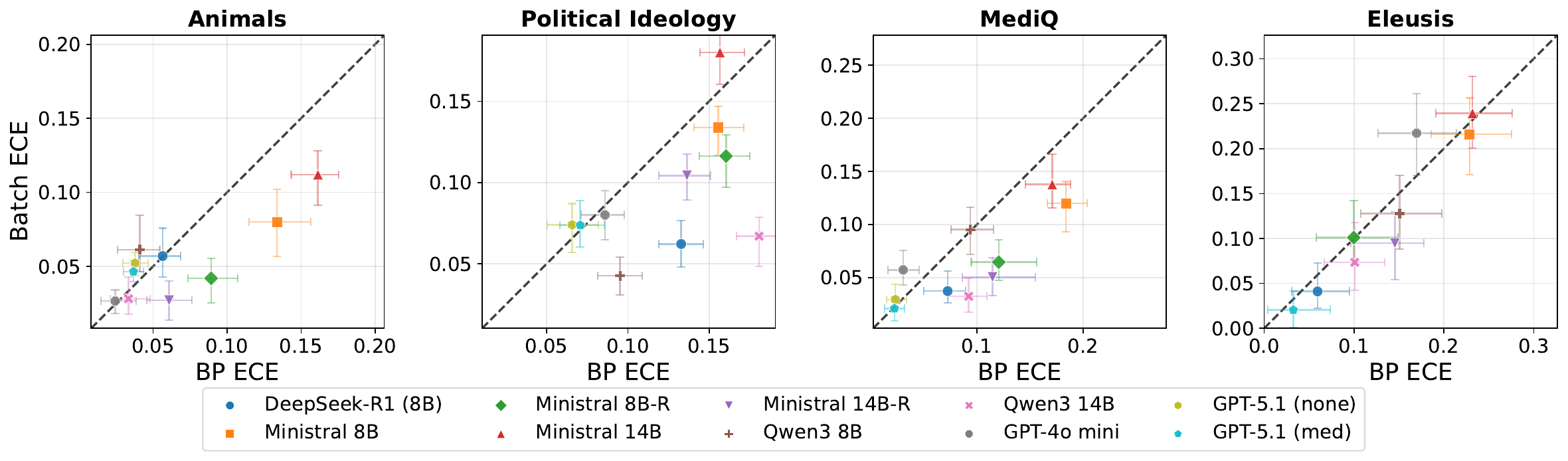}
    \caption{Comparing the expected calibration error (ECE) of the LLMs in batch mode vs.\ belief propagation mode.}
    \label{fig:batch_vs_seq_ece}
\end{figure}

In \Cref{fig:batch_vs_seq_ece} we compare ECE in batch and BP mode, as discussed in \Cref{sec:task_performance}. We find that for Animals, Political Ideology, and MediQ, most models are better calibrated in batch mode. We see no significant difference in calibration on Eleusis.

\subsection{When the elicited likelihood deviates from the oracle, task performance suffers}\label{app:oracle_auroc}
\begin{figure}[H]
    \centering
    \includegraphics[width=0.99\linewidth]{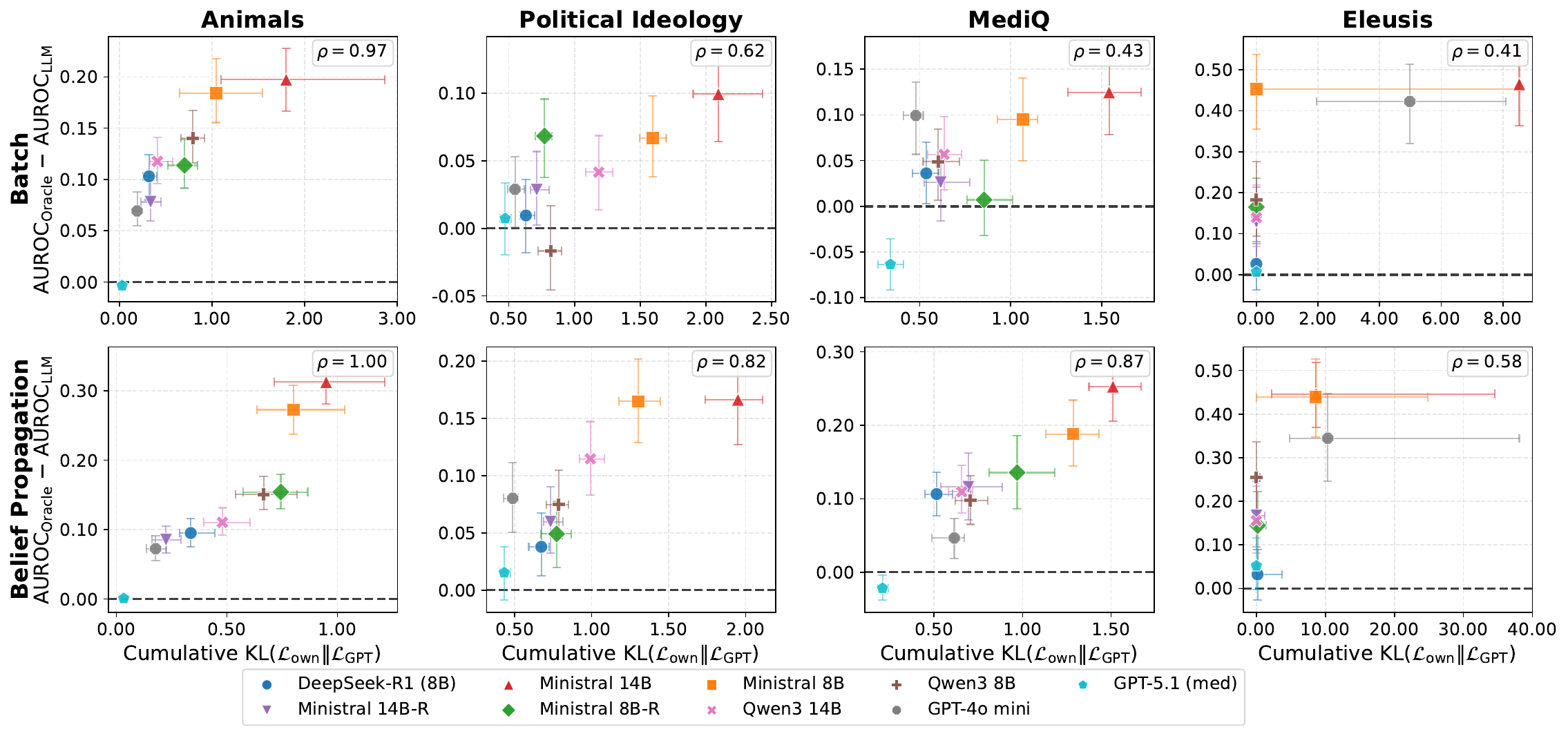}
    \caption{Relationship between likelihood divergence w.r.t.\ the oracle and task performance.
    Points above the dashed zero line indicate models whose posteriors underperform w.r.t.\ $p^\text{\tiny{or}}$.
    }
    \label{fig:oracle_auroc}
\end{figure}

In \Cref{fig:oracle_auroc}, we compare the task performance of $p^\text{\tiny{or}}$ (AUROC$_{\text{oracle}}$) and $q$ (AUROC$_{\text{LLM}}$) as a function of $\DKL(\ell || \ell^\text{\tiny{or}})$. 
We find that models whose elicited likelihoods substantially deviate from the oracle gain most in performance from $\ell^\text{\tiny{or}}$ substitution. 
However, for models whose elicited likelihood is close to the oracle, oracle substitution provides little performance gains and can even hurt performance. 
The correlations between the cumulative likelihood divergence and the oracle AUROC are systematically higher in the belief propagation mode than in the batch mode (consider, 0.3 gap between the two modes for MediQ, for instance). 

\section{KL vs.\ LER plots by evidence step}
\label{app:kl-ler-step-grid-plots}

In this section we provide the KL vs.\ LER plots for evidence steps 2--11 (or the maximum evidence step in the dataset) for all models and datasets, for both batch and belief propagation mode. The main takeaway is that we see similar behaviors across most evidence steps withing each model and dataset  .

\begin{figure}[ht]
    \centering
    \includegraphics[width=0.99\linewidth,height=0.9\textheight,keepaspectratio]{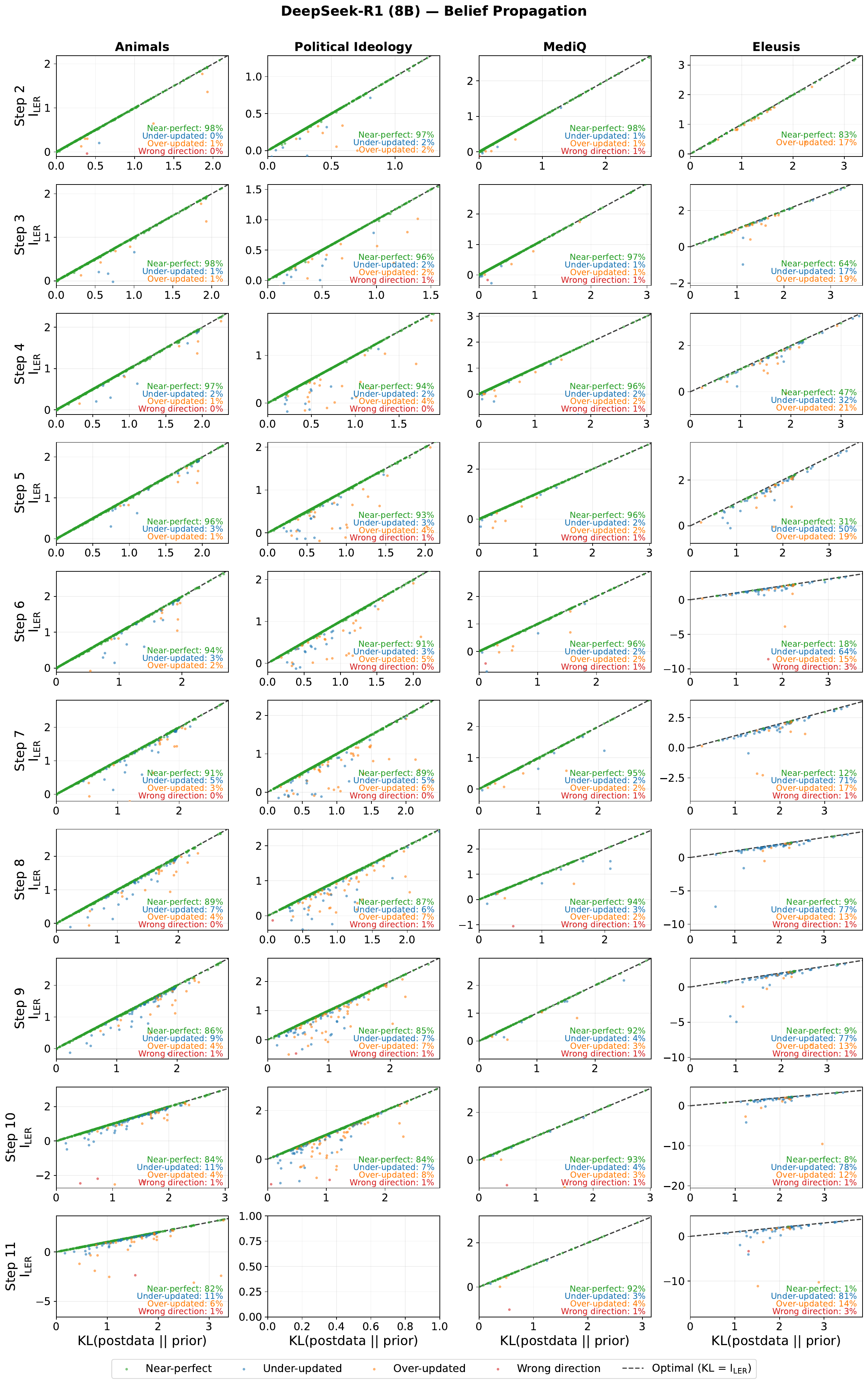}
    \caption{KL vs. LER for evidence steps 2--11 in belief propagation mode for DeepSeek-R1 distilled into Qwen3 8B.}
    \label{fig:kl_vs_ler_steps_sequential_deepseekr10528qwen38b}
\end{figure}

\begin{figure}[ht]
    \centering
    \includegraphics[width=0.99\linewidth,height=0.9\textheight,keepaspectratio]{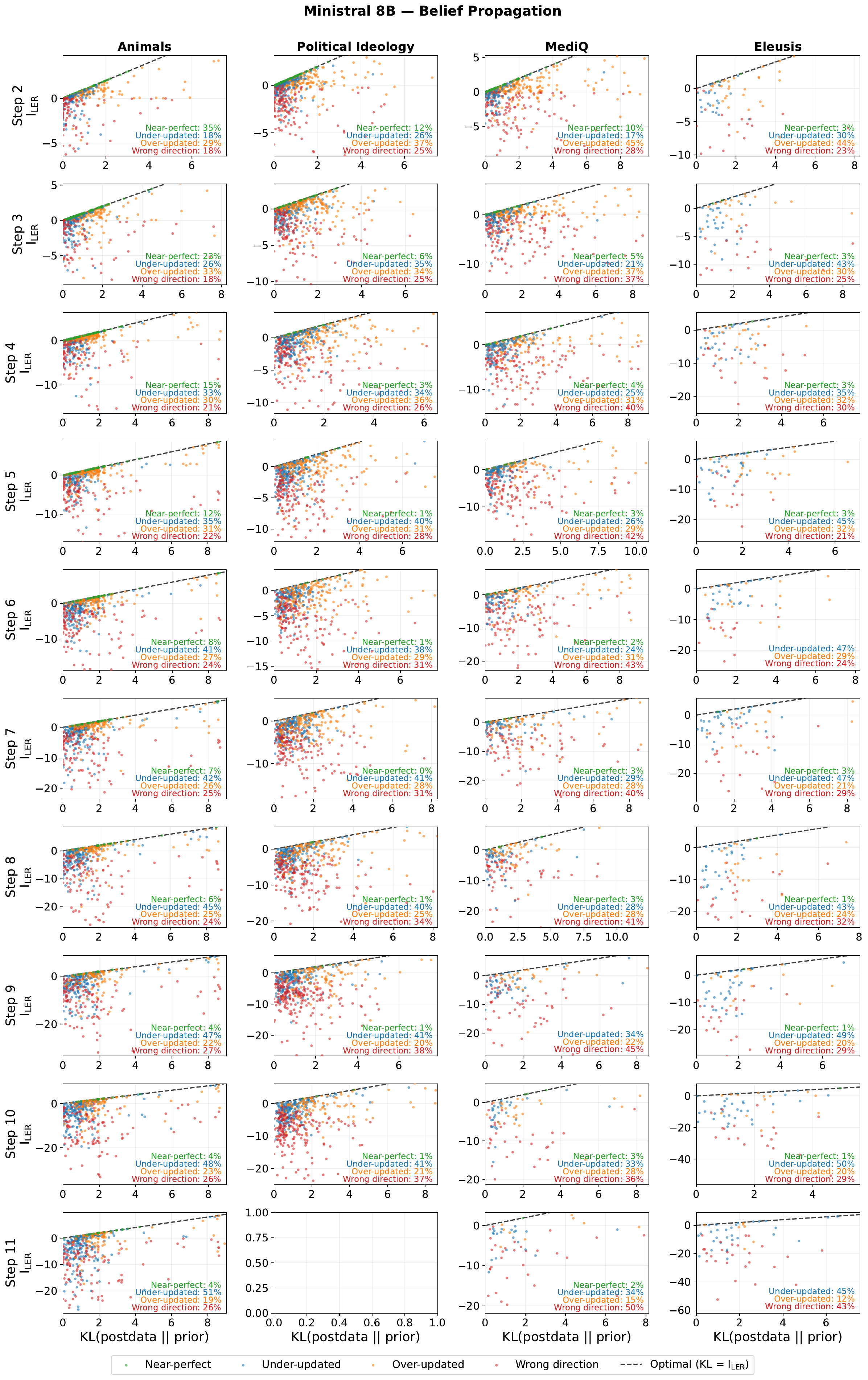}
    \caption{KL vs. LER for evidence steps 2--11 in belief propagation mode for Ministral 3 8B instruct.}
    \label{fig:kl_vs_ler_steps_sequential_ministral38binstruct2512}
\end{figure}

\begin{figure}[ht]
    \centering
    \includegraphics[width=0.99\linewidth,height=0.9\textheight,keepaspectratio]{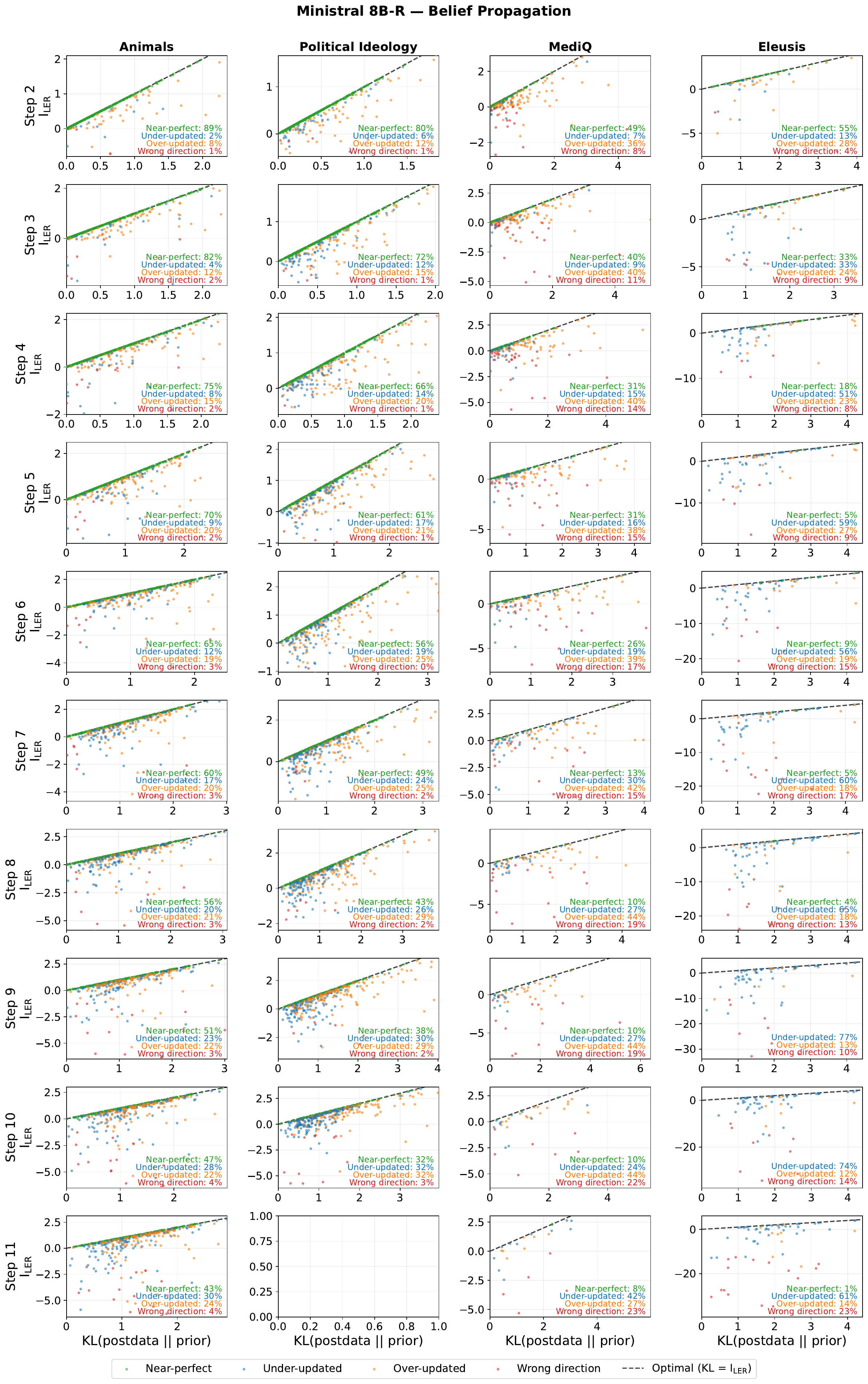}
    \caption{KL vs. LER for evidence steps 2--11 in belief propagation mode for Ministral 3 8B with reasoning.}
    \label{fig:kl_vs_ler_steps_sequential_ministral38breasoning2512}
\end{figure}

\begin{figure}[ht]
    \centering
    \includegraphics[width=0.99\linewidth,height=0.9\textheight,keepaspectratio]{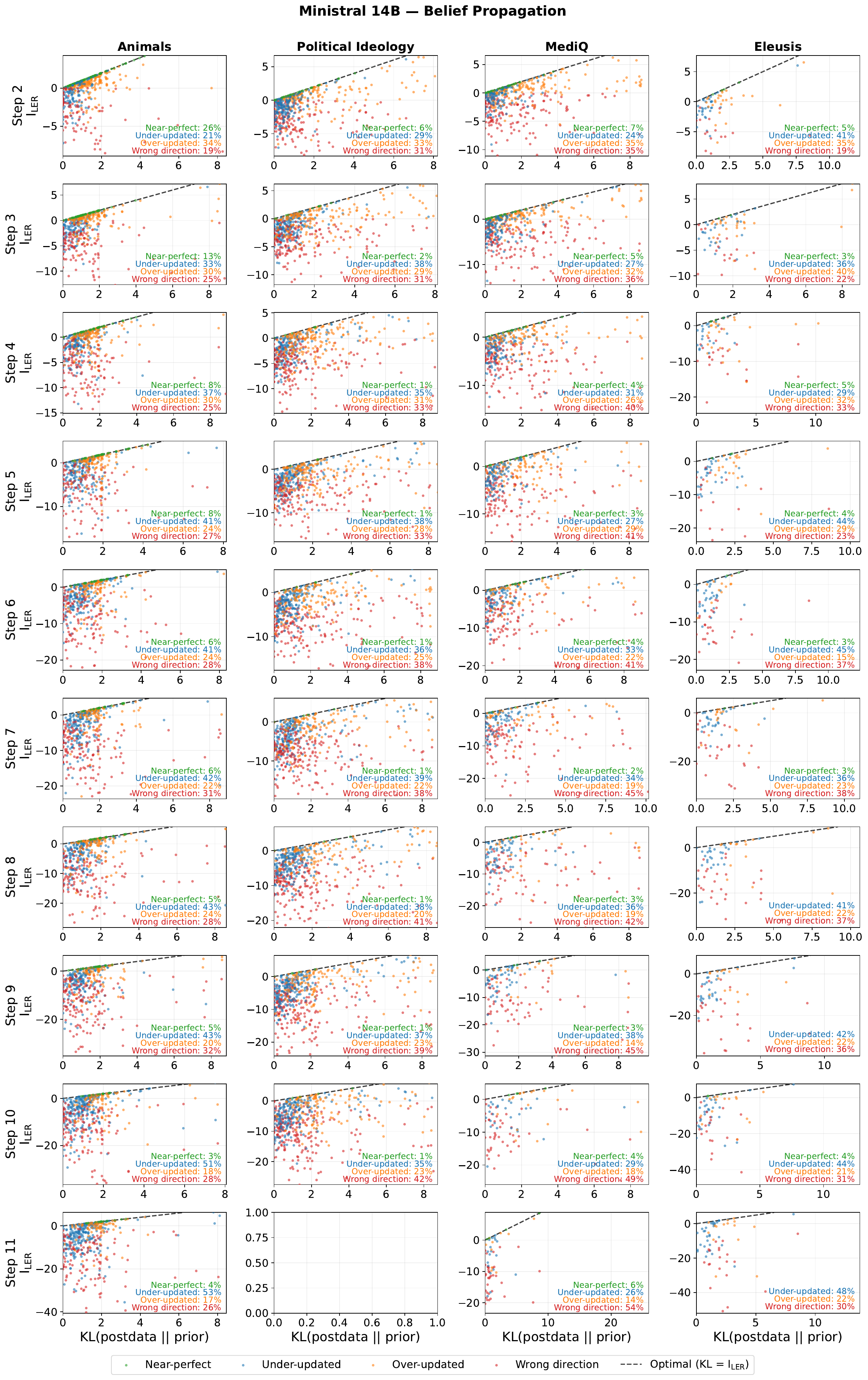}
    \caption{KL vs. LER for evidence steps 2--11 in belief propagation mode for Ministral 3 14B instruct.}
    \label{fig:kl_vs_ler_steps_sequential_ministral314binstruct2512}
\end{figure}

\begin{figure}[ht]
    \centering
    \includegraphics[width=0.99\linewidth,height=0.9\textheight,keepaspectratio]{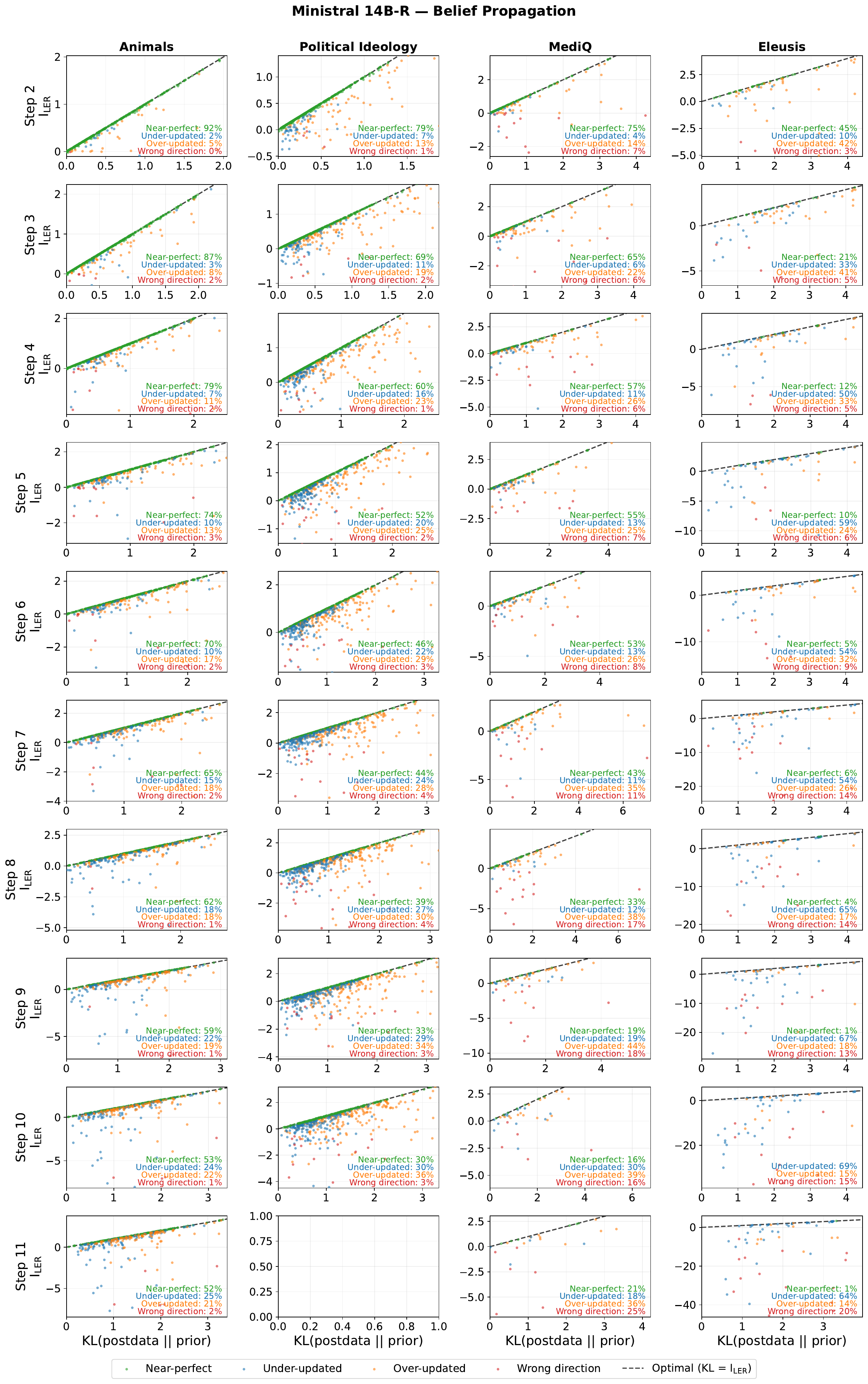}
    \caption{KL vs. LER for evidence steps 2--11 in belief propagation mode for Ministral 3 14B with reasoning.}
    \label{fig:kl_vs_ler_steps_sequential_ministral314breasoning2512}
\end{figure}

\begin{figure}[ht]
    \centering
    \includegraphics[width=0.99\linewidth,height=0.9\textheight,keepaspectratio]{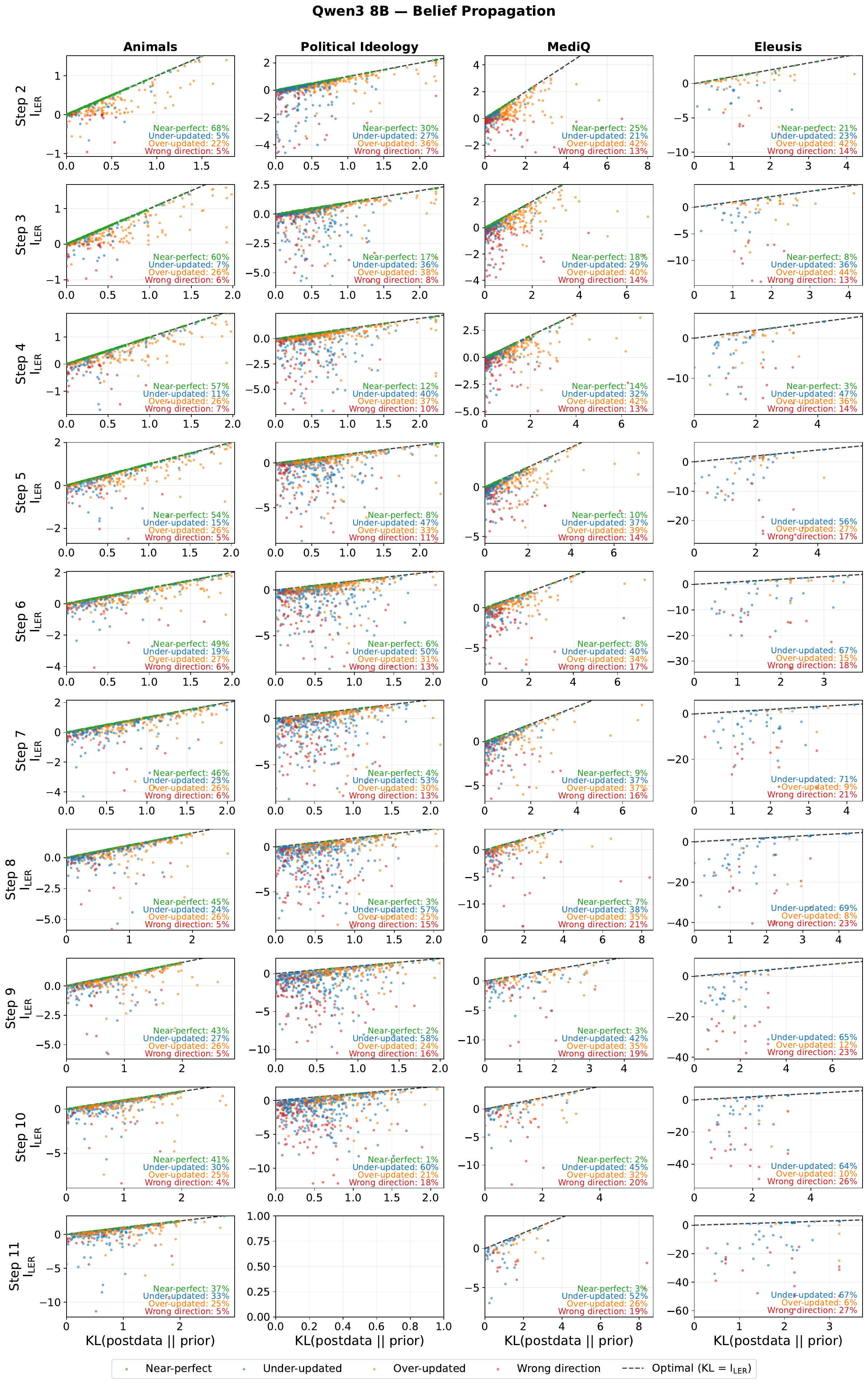}
    \caption{KL vs. LER for evidence steps 2--11 in belief propagation mode for Qwen3 8B.}
    \label{fig:kl_vs_ler_steps_sequential_qwen38b}
\end{figure}

\begin{figure}[ht]
    \centering
    \includegraphics[width=0.99\linewidth,height=0.9\textheight,keepaspectratio]{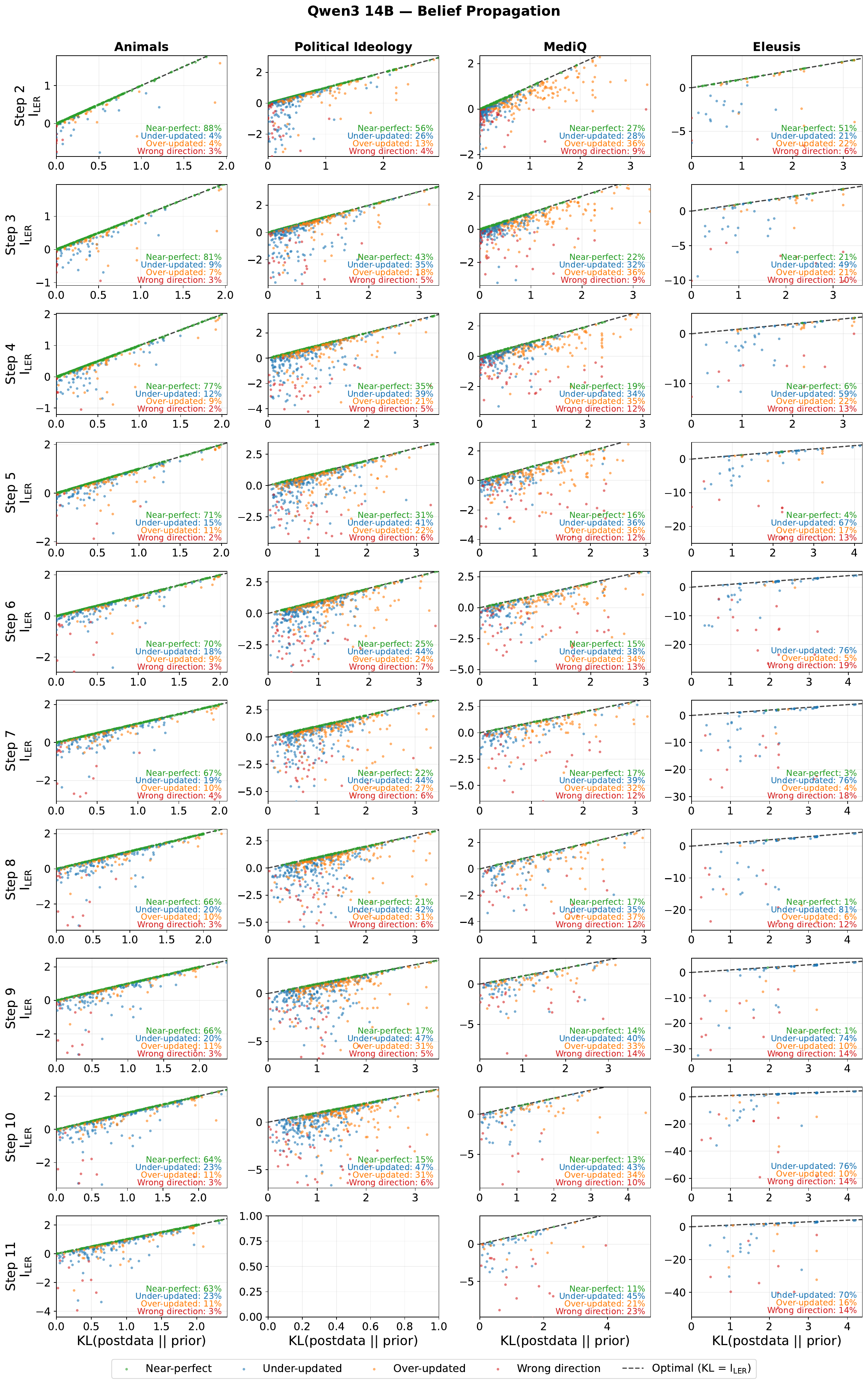}
    \caption{KL vs. LER for evidence steps 2--11 in belief propagation mode for Qwen3 14B.}
    \label{fig:kl_vs_ler_steps_sequential_qwen314b}
\end{figure}

\begin{figure}[ht]
    \centering
    \includegraphics[width=0.99\linewidth,height=0.9\textheight,keepaspectratio]{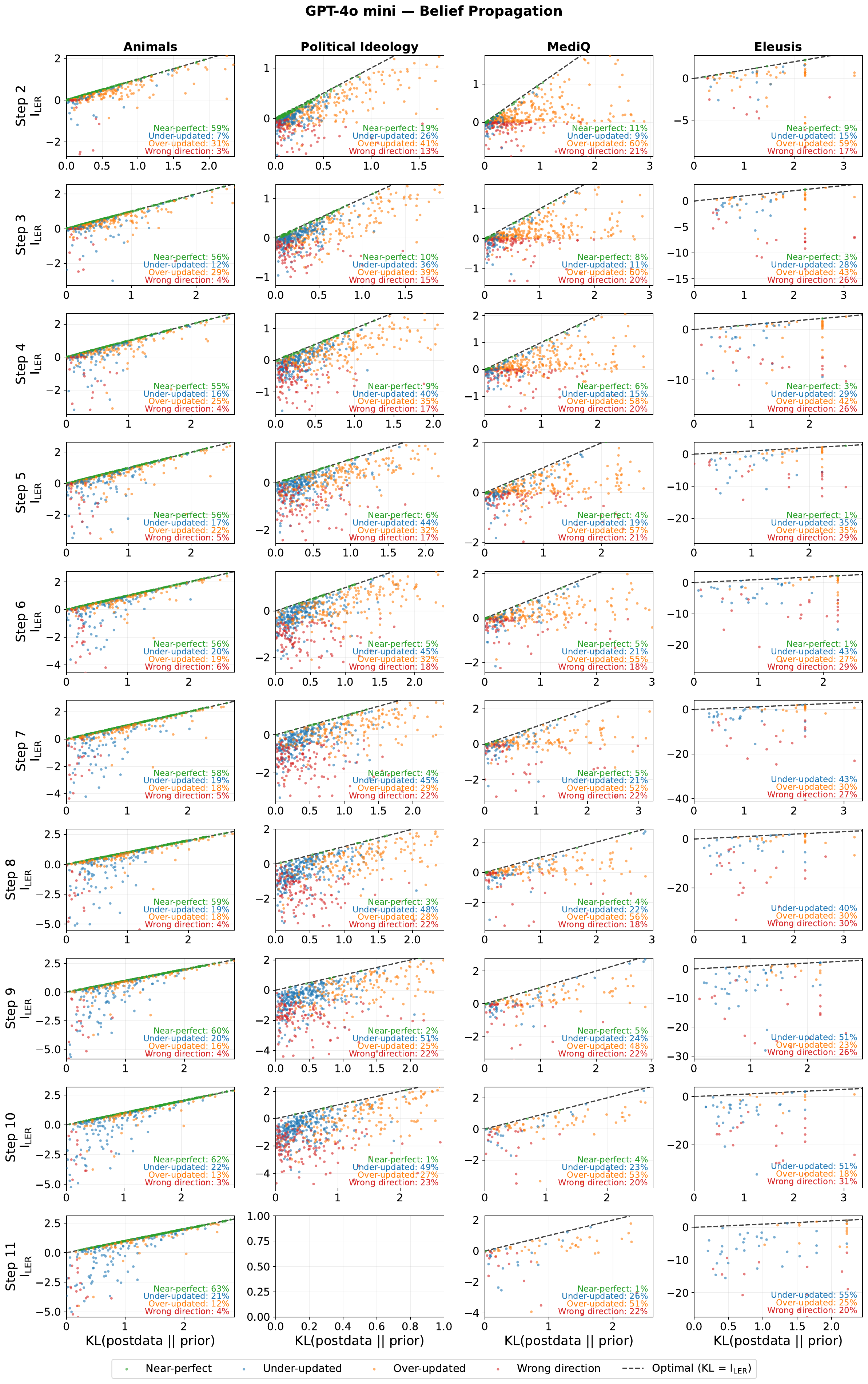}
    \caption{KL vs. LER for evidence steps 2--11 in belief propagation mode for GPT-4o mini.}
    \label{fig:kl_vs_ler_steps_sequential_gpt4omini}
\end{figure}

\begin{figure}[ht]
    \centering
    \includegraphics[width=0.99\linewidth,height=0.9\textheight,keepaspectratio]{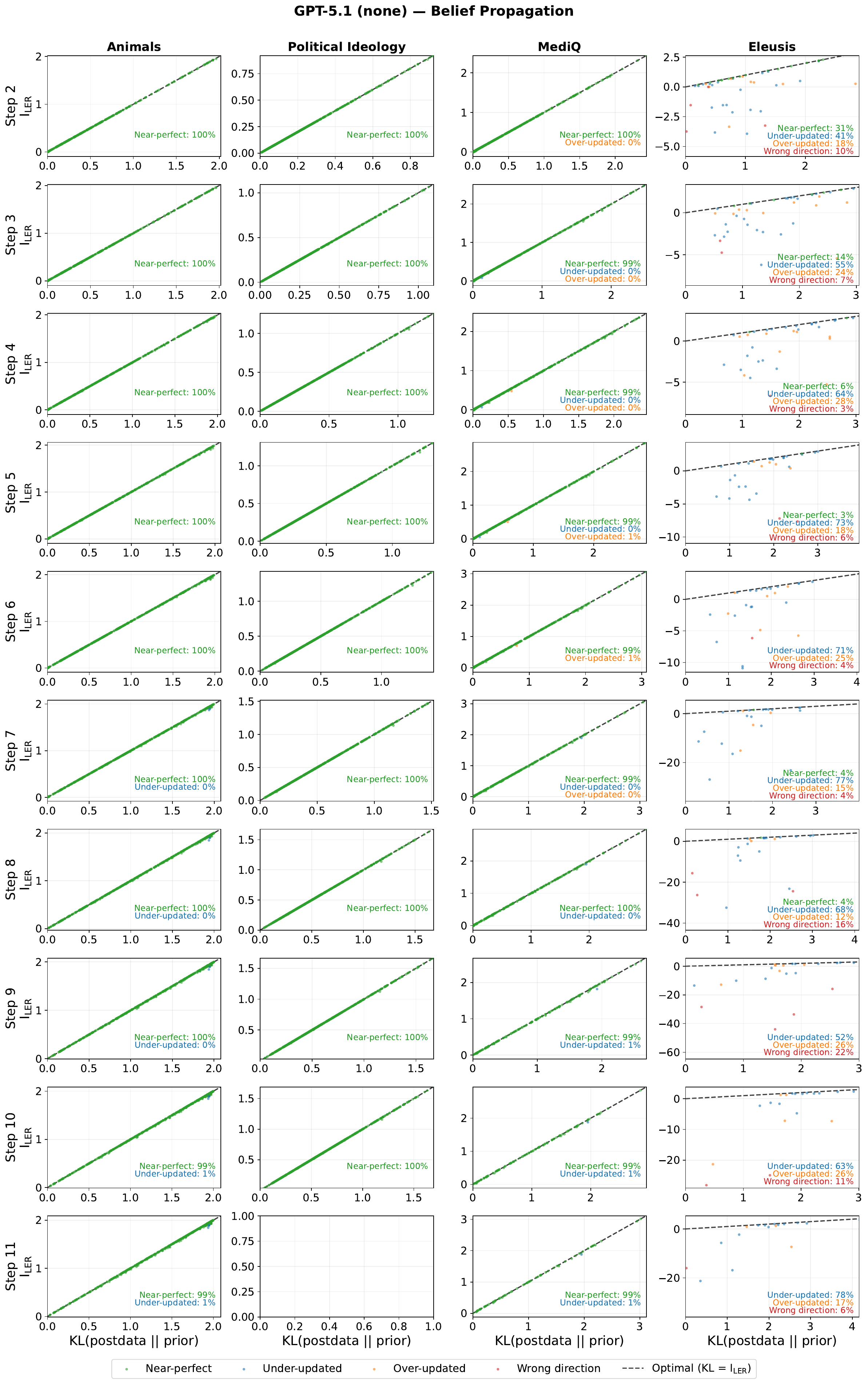}
    \caption{KL vs. LER for evidence steps 2--11 in belief propagation mode for GPT-5.1 without reasoning.}
    \label{fig:kl_vs_ler_steps_sequential_gpt51}
\end{figure}

\begin{figure}[ht]
    \centering
    \includegraphics[width=0.99\linewidth,height=0.9\textheight,keepaspectratio]{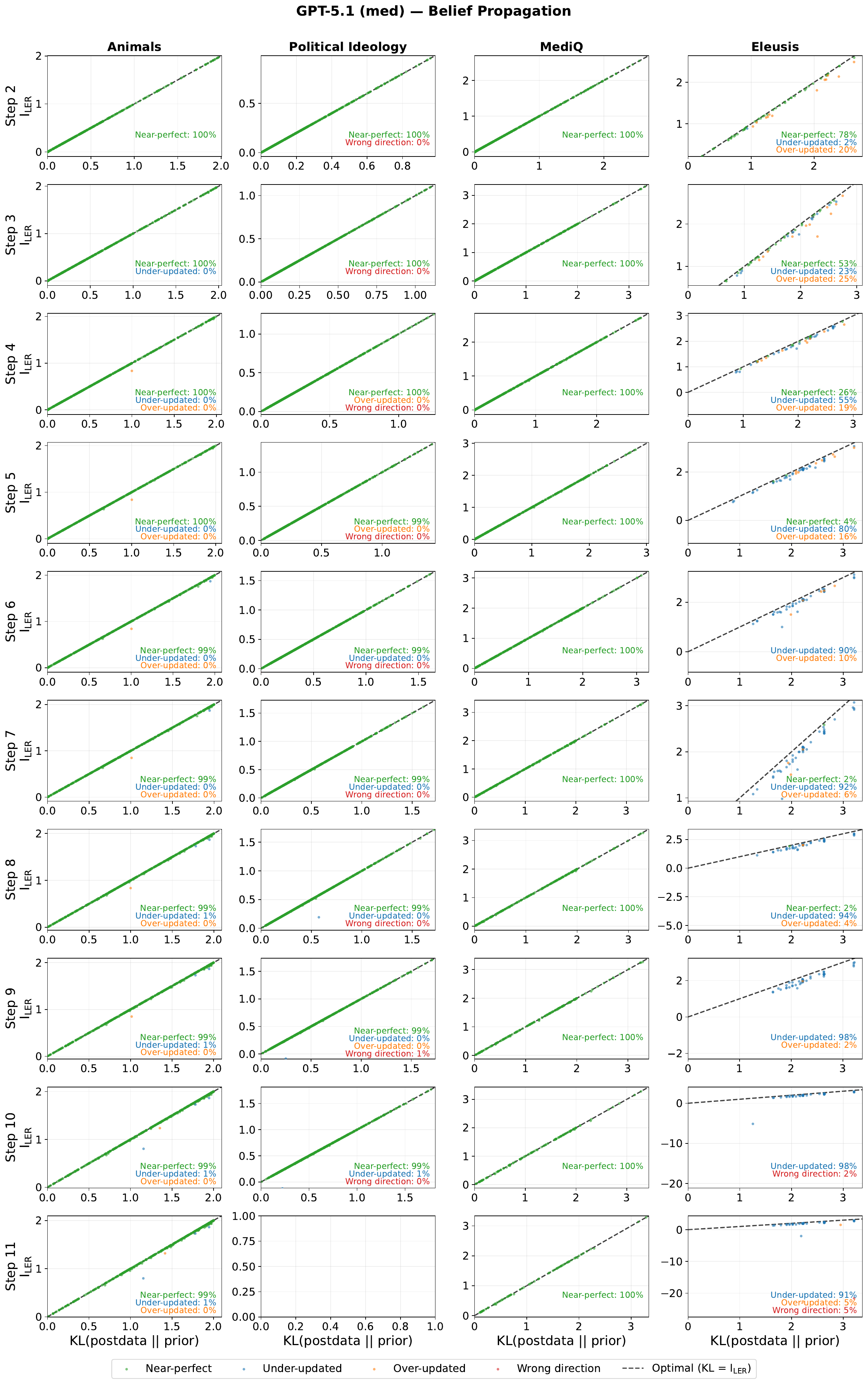}
    \caption{KL vs. LER for evidence steps 2--11 in belief propagation mode for GPT-5.1 with medium reasoning.}
    \label{fig:kl_vs_ler_steps_sequential_gpt51reasoningmedium}
\end{figure}

\begin{figure}[ht]
    \centering
    \includegraphics[width=0.99\linewidth,height=0.9\textheight,keepaspectratio]{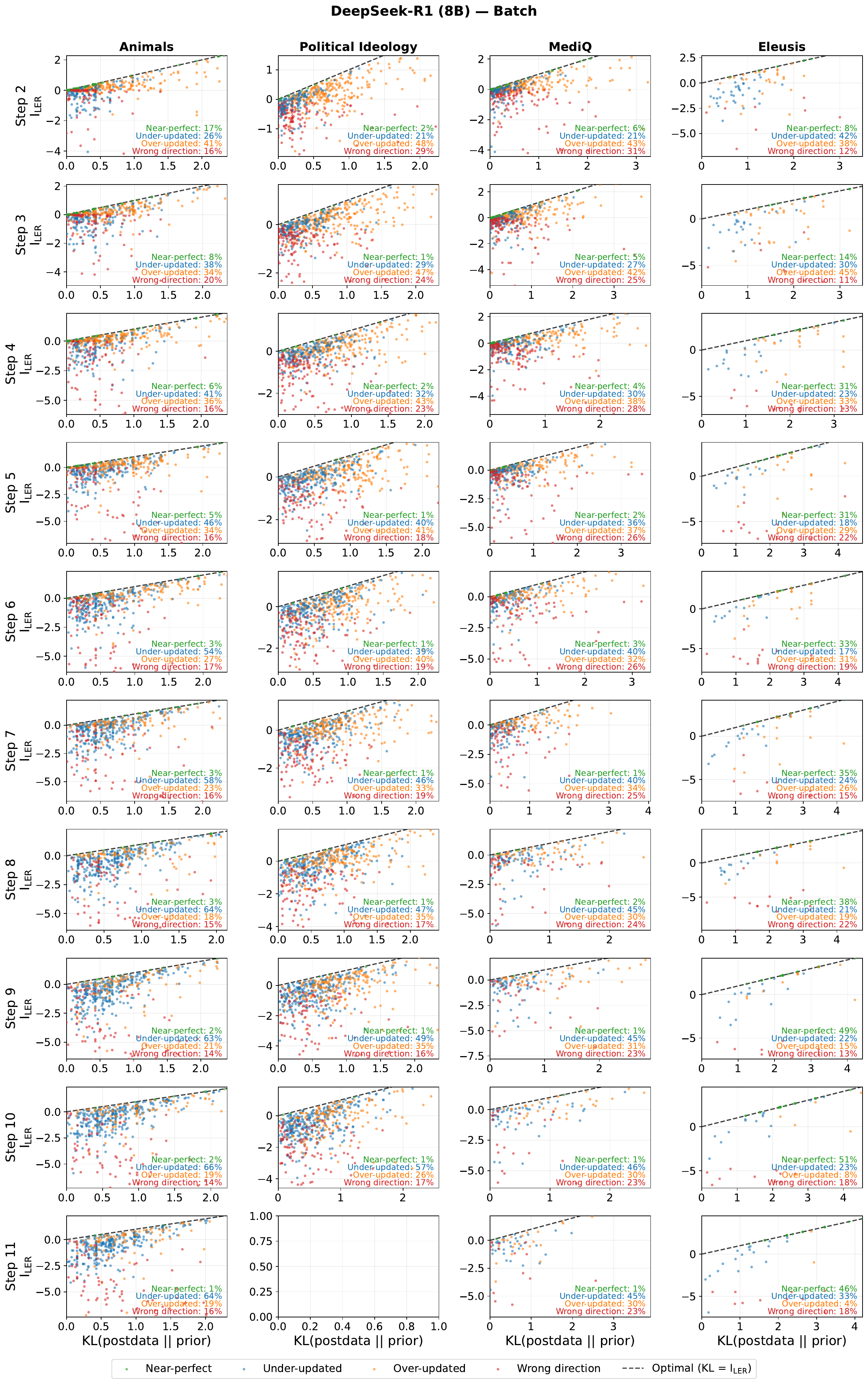}
    \caption{KL vs. LER for evidence steps 2--11 in batch mode for DeepSeek-R1 distilled into Qwen3 8B.}
    \label{fig:kl_vs_ler_steps_batch_deepseekr10528qwen38b}
\end{figure}

\begin{figure}[ht]
    \centering
    \includegraphics[width=0.99\linewidth,height=0.9\textheight,keepaspectratio]{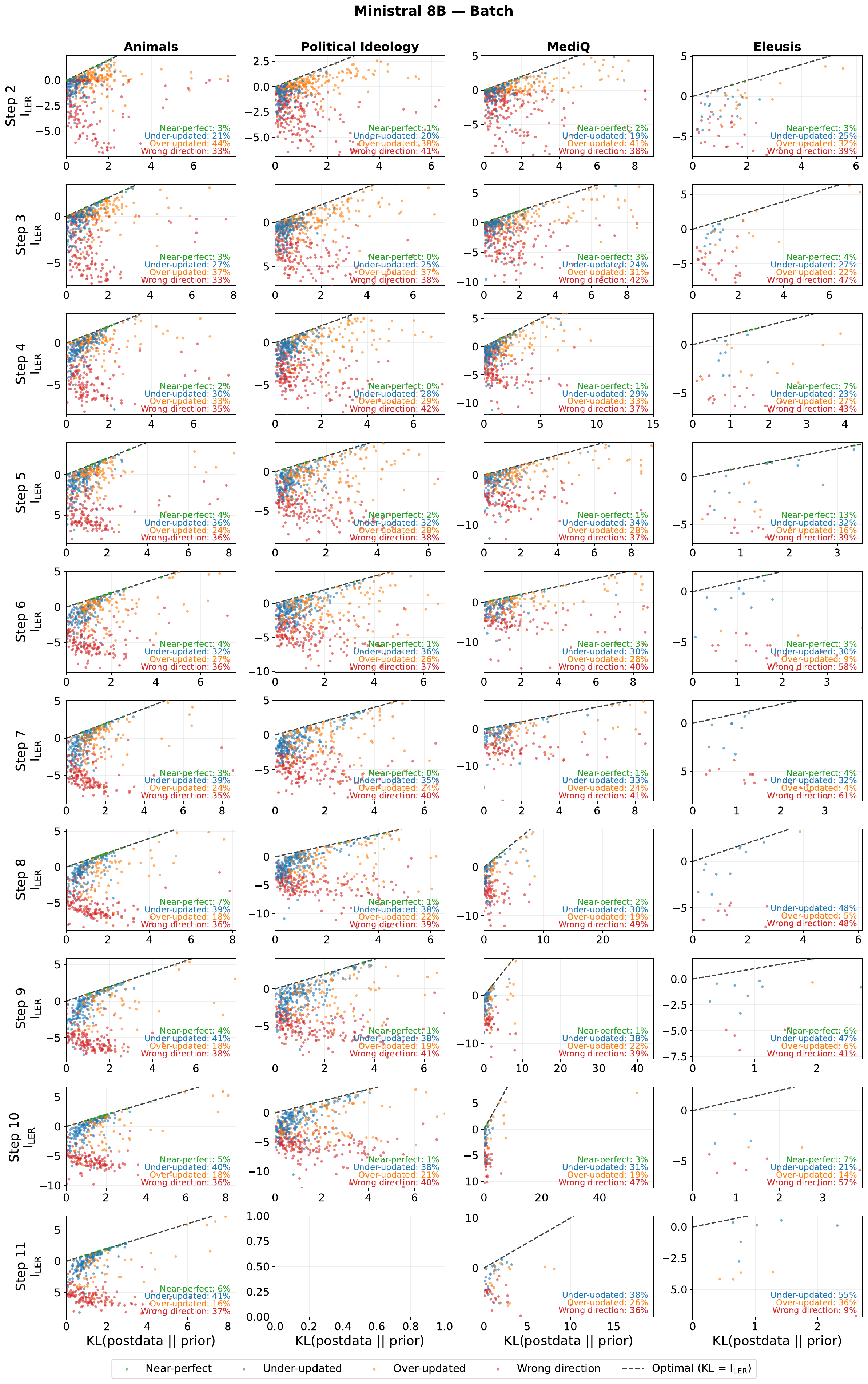}
    \caption{KL vs. LER for evidence steps 2--11 in batch mode for Ministral 3 8B instruct.}
    \label{fig:kl_vs_ler_steps_batch_ministral38binstruct2512}
\end{figure}

\begin{figure}[ht]
    \centering
    \includegraphics[width=0.99\linewidth,height=0.9\textheight,keepaspectratio]{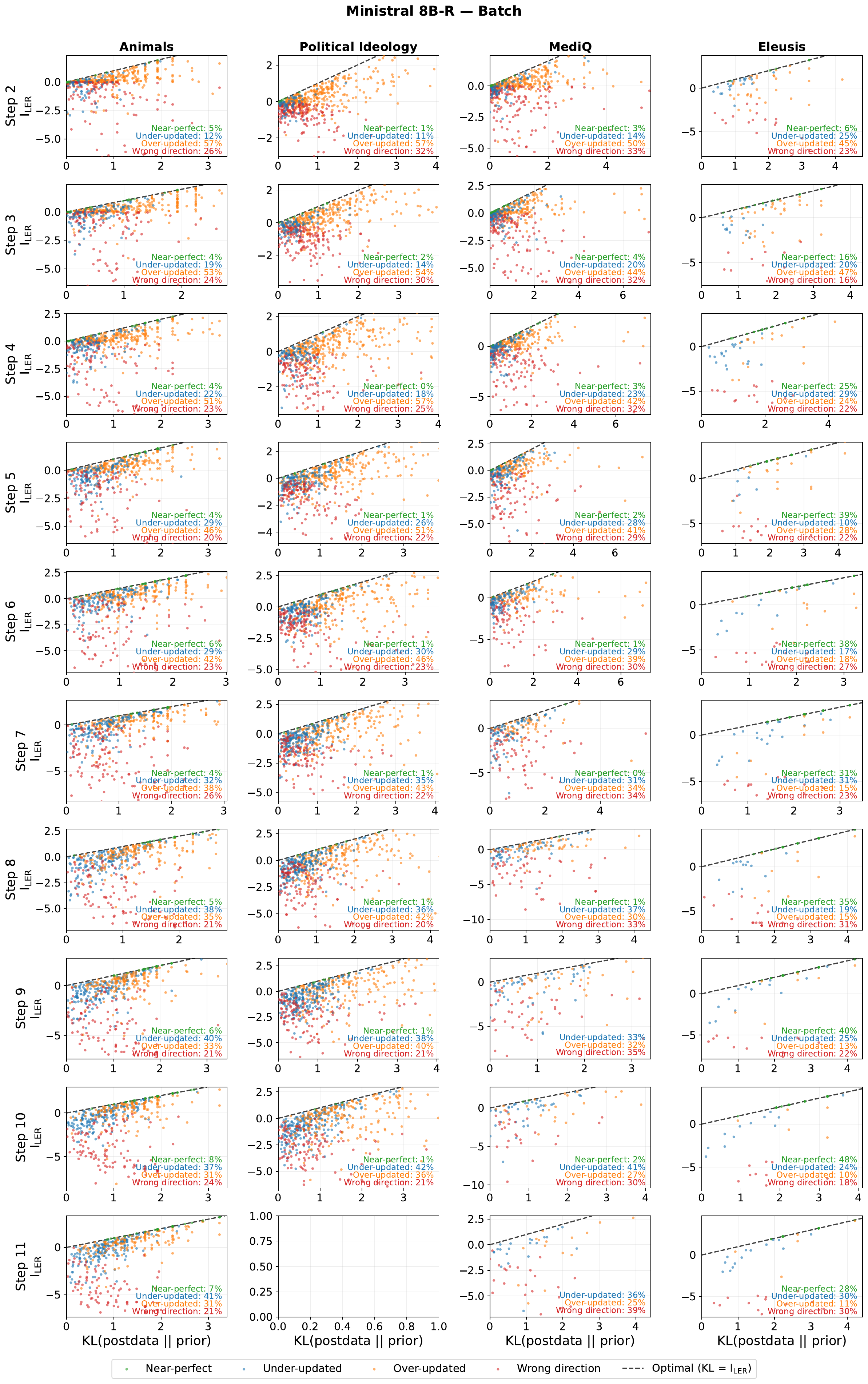}
    \caption{KL vs. LER for evidence steps 2--11 in batch mode for Ministral 3 8B with reasoning.}
    \label{fig:kl_vs_ler_steps_batch_ministral38breasoning2512}
\end{figure}

\begin{figure}[ht]
    \centering
    \includegraphics[width=0.99\linewidth,height=0.9\textheight,keepaspectratio]{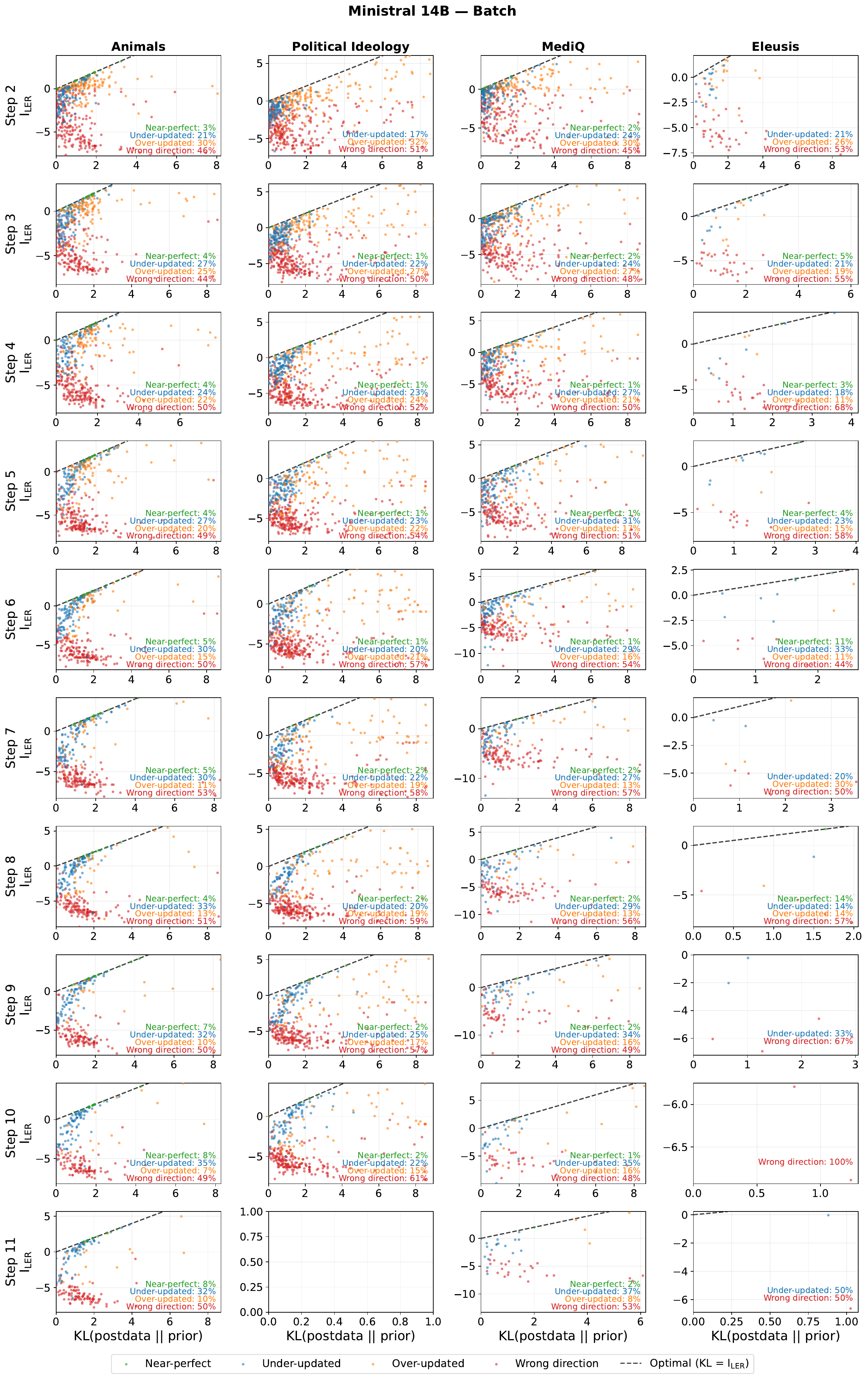}
    \caption{KL vs. LER for evidence steps 2--11 in batch mode for Ministral 3 14B instruct.}
    \label{fig:kl_vs_ler_steps_batch_ministral314binstruct2512}
\end{figure}

\begin{figure}[ht]
    \centering
    \includegraphics[width=0.99\linewidth,height=0.9\textheight,keepaspectratio]{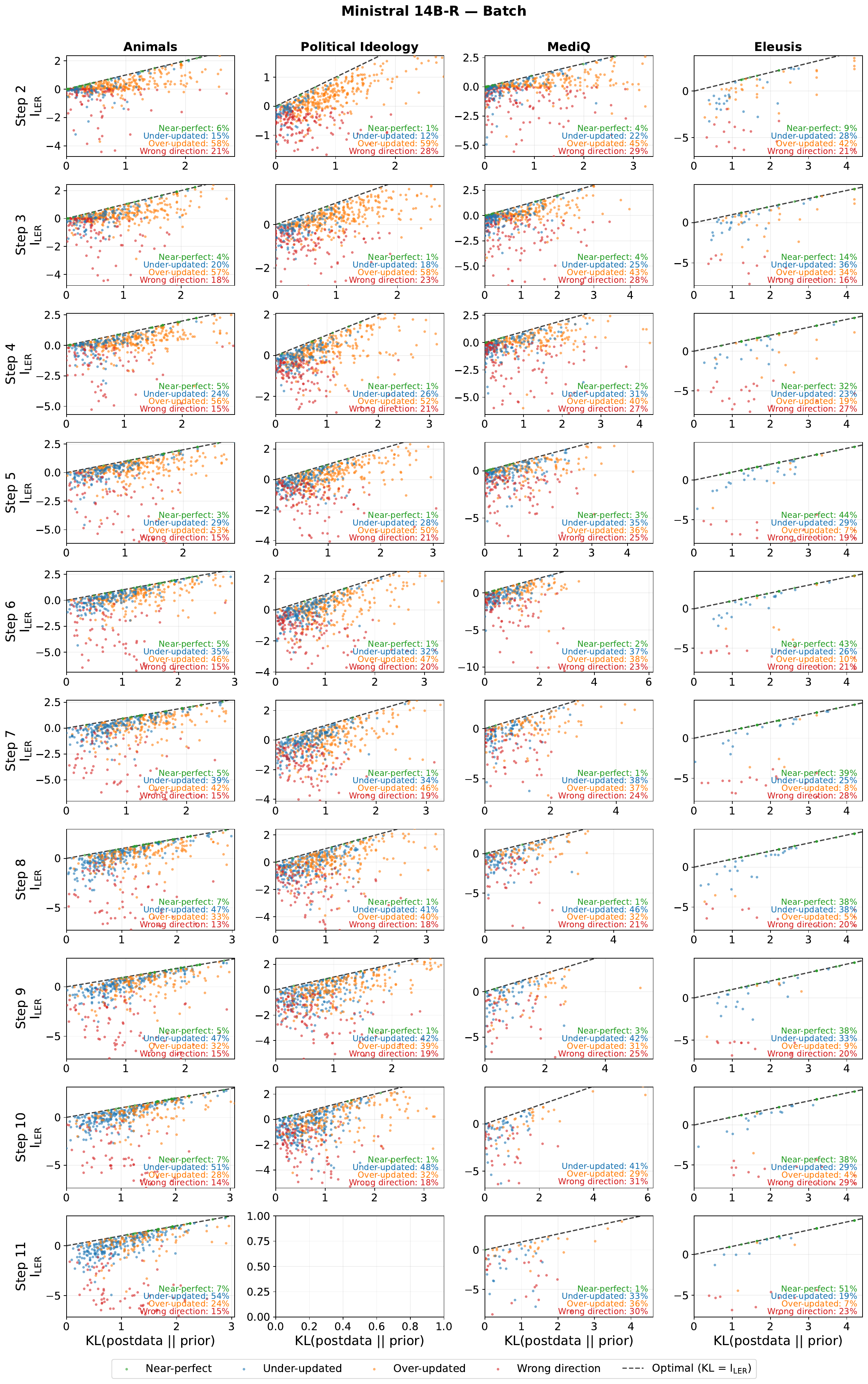}
    \caption{KL vs. LER for evidence steps 2--11 in batch mode for Ministral 3 14B with reasoning.}
    \label{fig:kl_vs_ler_steps_batch_ministral314breasoning2512}
\end{figure}

\begin{figure}[ht]
    \centering
    \includegraphics[width=0.99\linewidth,height=0.9\textheight,keepaspectratio]{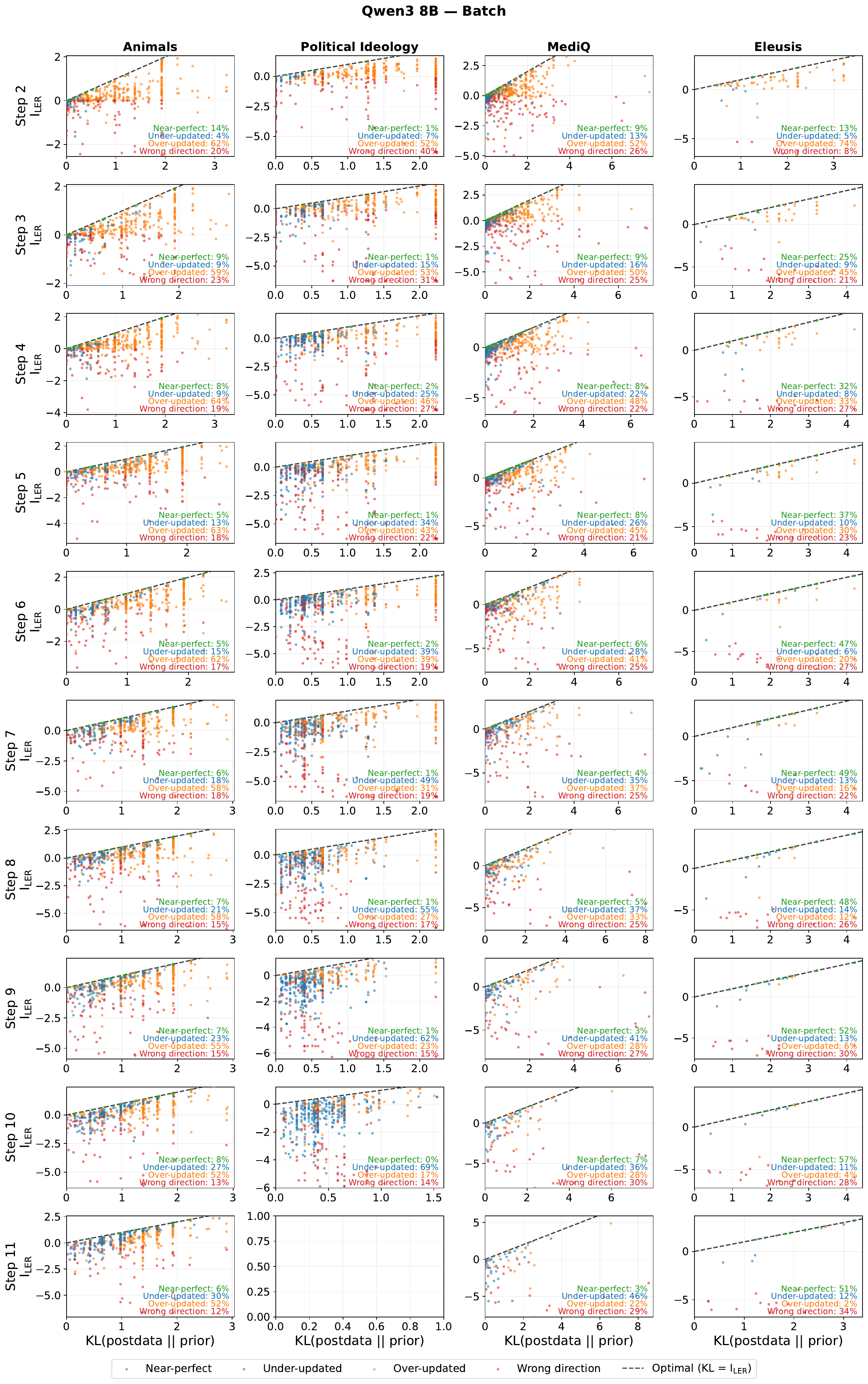}
    \caption{KL vs. LER for evidence steps 2--11 in batch mode for Qwen3 8B.}
    \label{fig:kl_vs_ler_steps_batch_qwen38b}
\end{figure}

\begin{figure}[ht]
    \centering
    \includegraphics[width=0.99\linewidth,height=0.9\textheight,keepaspectratio]{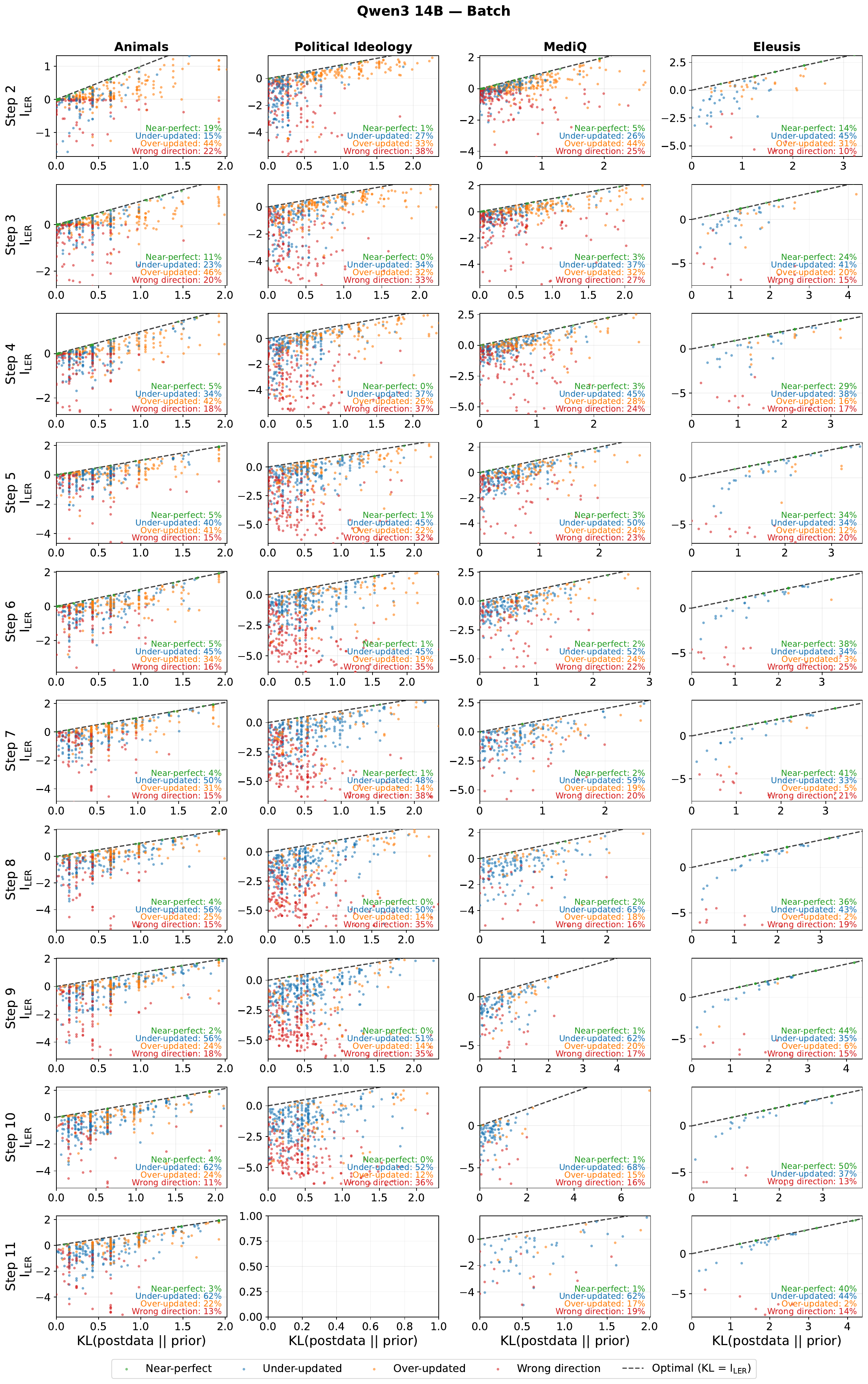}
    \caption{KL vs. LER for evidence steps 2--11 in batch mode for Qwen3 14B.}
    \label{fig:kl_vs_ler_steps_batch_qwen314b}
\end{figure}

\begin{figure}[ht]
    \centering
    \includegraphics[width=0.99\linewidth,height=0.9\textheight,keepaspectratio]{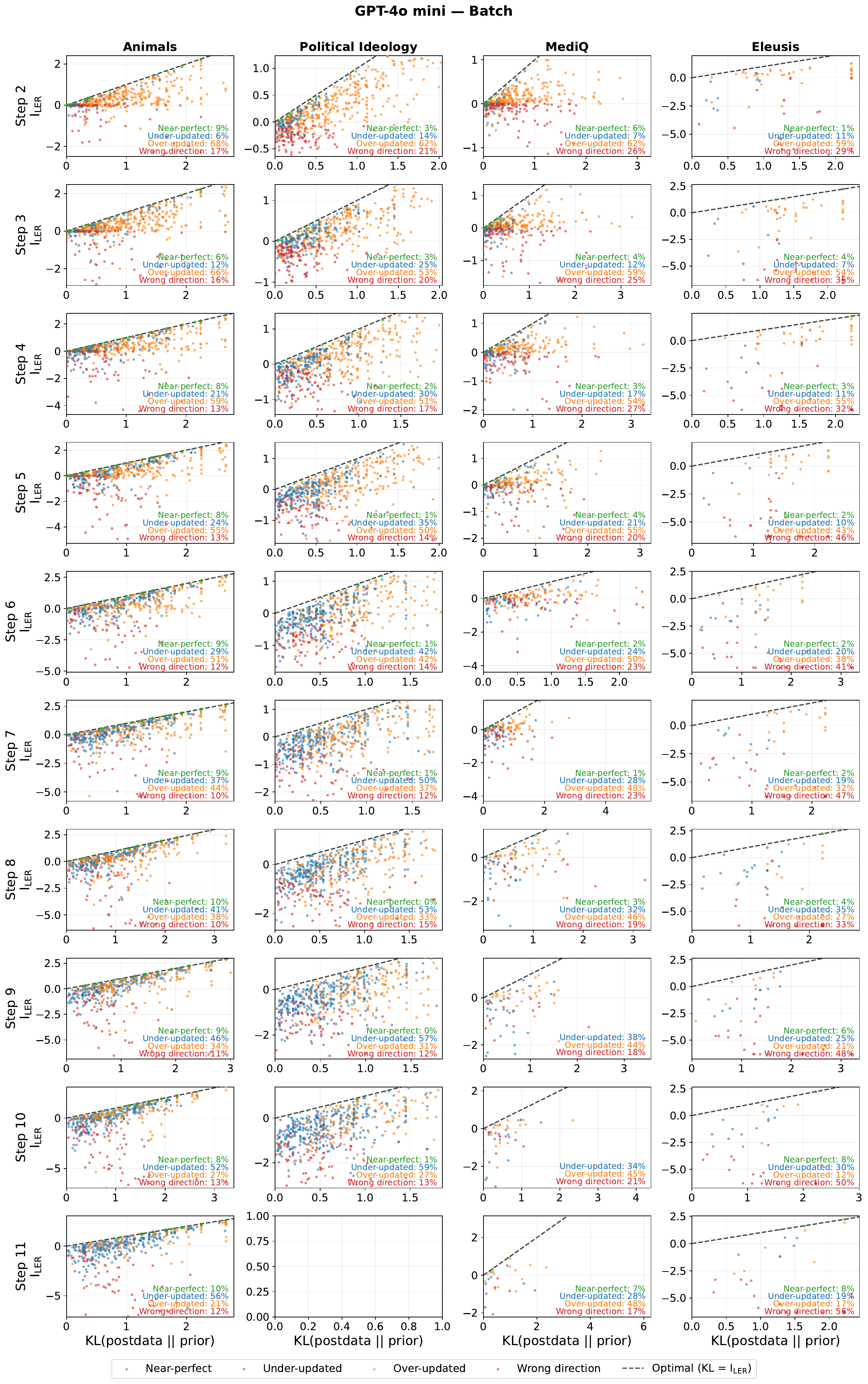}
    \caption{KL vs. LER for evidence steps 2--11 in batch mode for GPT-4o mini.}
    \label{fig:kl_vs_ler_steps_batch_gpt4omini}
\end{figure}

\begin{figure}[ht]
    \centering
    \includegraphics[width=0.99\linewidth,height=0.9\textheight,keepaspectratio]{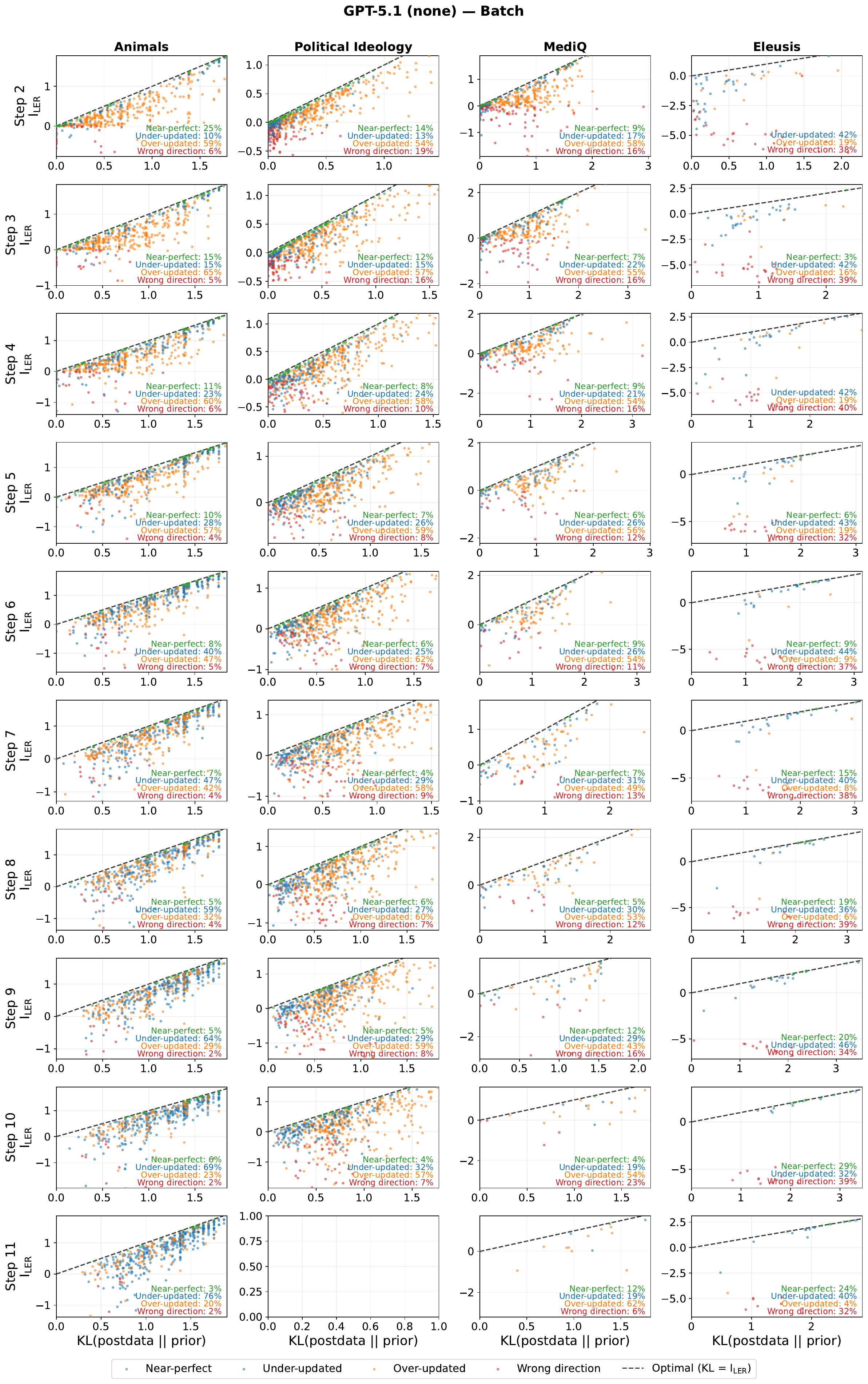}
    \caption{KL vs. LER for evidence steps 2--11 in batch mode for GPT-5.1 without reasoning.}
    \label{fig:kl_vs_ler_steps_batch_gpt51}
\end{figure}

\begin{figure}[ht]
    \centering
    \includegraphics[width=0.99\linewidth,height=0.9\textheight,keepaspectratio]{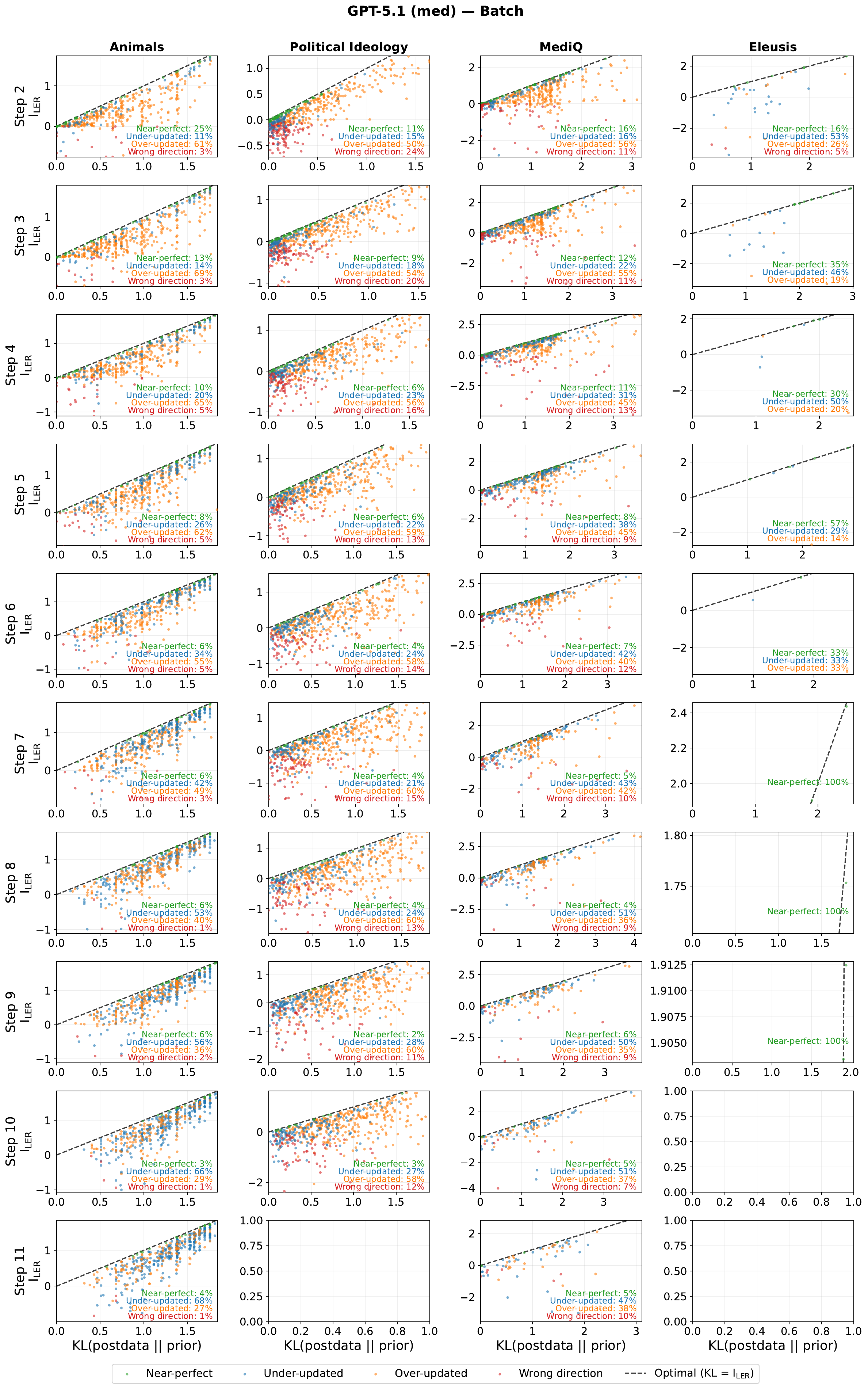}
    \caption{KL vs. LER for evidence steps 2--11 in batch mode for GPT-5.1 with medium reasoning.}
    \label{fig:kl_vs_ler_steps_batch_gpt51reasoningmedium}
\end{figure}

\Cref{fig:kl_vs_ler_steps_sequential_deepseekr10528qwen38b} to \Cref{fig:kl_vs_ler_steps_sequential_gpt51reasoningmedium} show the results for belief propagation mode.
Similarly, \Cref{fig:kl_vs_ler_steps_batch_deepseekr10528qwen38b} to \Cref{fig:kl_vs_ler_steps_batch_gpt51reasoningmedium} show the KL vs. LER plots for all models and datasets in batch mode.

\end{document}